\pdfoutput=1
\documentclass[10pt,twocolumn,letterpaper]{article}

\usepackage{text}
\usepackage[table]{xcolor}
\usepackage{wrapfig}
\usepackage{placeins}
\providecommand{\keywords}[1]{\par\noindent\textbf{Keywords:}\enspace #1}
\newcommand{\kwsep}{\unskip\space\textperiodcentered\space}

\usepackage{soul}
\usepackage{wrapfig}
\usepackage{graphicx}
\usepackage{booktabs}
\usepackage{xcolor}
\usepackage{colortbl,multirow}

\definecolor{Palette5}{RGB}{244, 233, 183}
\definecolor{OursColor}{RGB}{227, 220, 234}
\definecolor{CatToM}{RGB}{255,200,200}
\definecolor{CatAll}{RGB}{200,220,255}
\definecolor{CatOther}{RGB}{255,235,180}
\definecolor{cvprblue}{rgb}{0.21,0.49,0.74}

\usepackage[pagebackref,breaklinks,colorlinks,allcolors=cvprblue]{hyperref}

\newcommand\blfootnote[1]{%
  \begingroup
  \renewcommand\thefootnote{}\footnote{#1}%
  \addtocounter{footnote}{-1}%
  \endgroup
}

\newcommand{\modelname}[1]{\textbf{SeeClear}}

\title{SeeClear: Reliable Transparent Object Depth Estimation via Generative Opacification}

\author{
Xiaoying Wang$^{1*}$ \;
Yumeng He$^{1,2*}$ \;
Jingkai Shi$^{1,2*}$ \;
Jiayin Lu$^{1}$ \;
Yin Yang$^{3}$ \;
Ying Jiang$^{1}$ \;
Chenfanfu Jiang$^{1}$
}

\begin{document}

\setcounter{footnote}{0}

\twocolumn[{%
\renewcommand\twocolumn[1][]{#1}%
\maketitle

\begin{center}
    \centering
    \captionsetup{type=figure}
    \includegraphics[width=\textwidth]{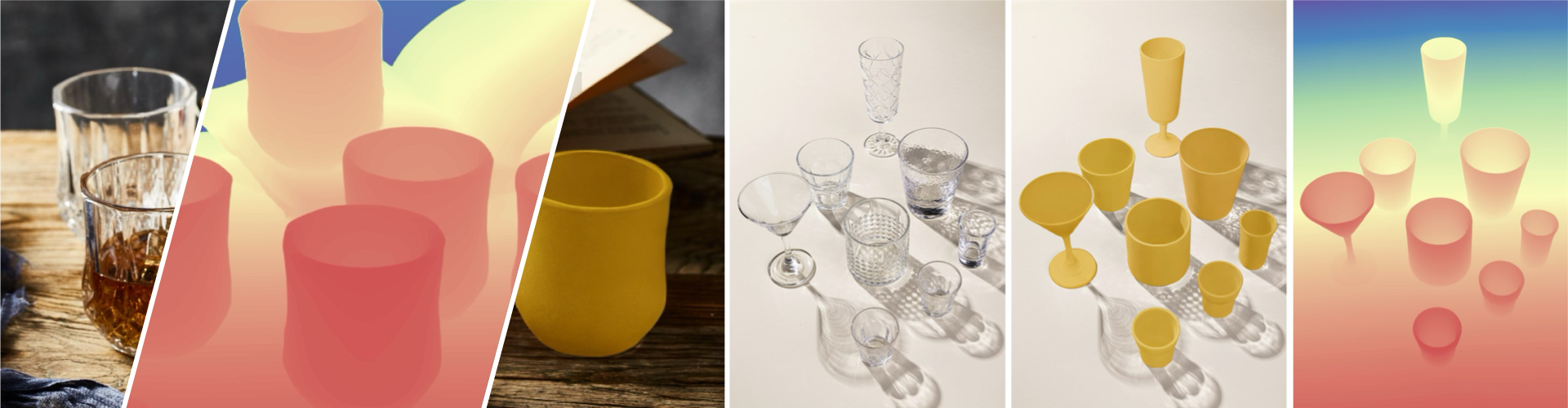}
    \captionof{figure}{\modelname{} is a novel framework that converts transparent objects into generative opaque images, predicting stable and accurate depth for transparent objects.}
    \label{fig:teaser}
\end{center}
}]

% TODO
\blfootnote{$^{*}$ Equal contribution. $^{1}$ University of California, Los Angeles. $^{2}$ University of Southern California. $^{3}$ University of Utah.
\url{xwang648@usc.edu}, \url{heyumeng@usc.edu}, \url{shikings@usc.edu}, \url{jiayin_lu@math.ucla.edu}, \url{yangzzzy@gmail.com}, \url{anajymua@gmail.com}, \url{cffjiang@ucla.edu}
}

\begin{abstract}

\vspace{-5pt}

Monocular depth estimation remains challenging for transparent objects, where refraction and transmission are difficult to model and break the appearance assumptions used by depth networks. As a result, state-of-the-art estimators often produce unstable or incorrect depth predictions for transparent materials. We propose \modelname{}, a novel framework that converts transparent objects into generative opaque images, enabling stable monocular depth estimation for transparent objects. Given an input image, we first localize transparent regions and transform their refractive appearance into geometrically consistent opaque shapes using a diffusion-based generative opacification module. The processed image is then fed into an off-the-shelf monocular depth estimator without retraining or architectural changes. To train the opacification model, we construct \textbf{SeeClear-396k}, a synthetic dataset containing 396k paired transparent-opaque renderings. Experiments on both synthetic and real-world datasets show that \modelname{} significantly improves depth estimation for transparent objects. Project page: \url{https://heyumeng.com/SeeClear-web/}

\end{abstract}

%%%%%%%%%%%%%%%%%%%%%%%%%%%%%%%%%%%%%%%%%%%%%%%%%%%%%%%%%%%%
% introduction
%%%%%%%%%%%%%%%%%%%%%%%%%%%%%%%%%%%%%%%%%%%%%%%%%%%%%%%%%%%%

\section{Introduction}
\label{sec:intro}
Monocular depth estimation underpins 3D reconstruction \cite{guo2025murre, newcombe2011kinectfusion}, navigation \cite{hodan2020bop}, and robotic manipulation \cite{shankar2022learned, schmidt2015depth, li2024visual}, and is a key enabler for grasping \cite{mahler2017dexnet20deeplearning, tenpas2017gpd}, collision avoidance \cite{flacco2012depth, flacco2015depth}, and real-to-simulation pipelines \cite{schonberger2016sfm, qiu2025joint3dpointcloud}. Though recent monocular depth models achieve strong performance \cite{zhou2017scene, silberman11indoor, Silberman:ECCV12}, they often fail in real-world scenes containing objects with complex appearance properties. In particular, transparent objects pose a severe challenge due to refractive and transmissive effects that violate common photometric assumptions \cite{ma2025physliquidphysicsinformeddatasetestimating}, leading to erroneous depth predictions that can propagate to downstream simulation and manipulation tasks. This discrepancy is particularly problematic in physical scene understanding. For example, inaccurate depth estimation of transparent cabinets can lead to incorrect reconstruction of collision regions.

Prior work addresses transparent depth estimation by fine-tuning depth models specifically for transparent objects \cite{tosi2024diffusion, costanzino2023learning, dkt2025}, achieving strong performance on transparent scenes. However, these approaches rely on large-scale paired transparent RGB–depth data and introduce distribution shifts. Moreover, as state-of-the-art depth models continue to evolve rapidly \cite{yang2024depthanythingv2}, every updated backbone requires additional fine-tuning. Although modern foundation depth models remain challenged by transparent depth estimation, they demonstrate strong generalization and high accuracy for opaque objects. This naturally raises the question: can we leverage transparent-object data without modifying or retraining the depth model, while remaining compatible with foundation depth models, to enable accurate depth estimation for transparent objects?

We propose \modelname{}, a plug-and-play front-end framework for accurate and stable monocular transparent-object depth estimation. The method first localizes transparent regions and converts them into geometrically consistent opaque appearances. Specifically, a segmentation model provides object masks as structural guidance, while a diffusion model generates diffuse textures to replace refractive appearances while preserving object boundaries and shading. The opacified image is then composited with the original background to form the final input for the depth estimator with a mask refinement module. To train the generative opacification module, we construct \textbf{SeeClear-396k}, a large-scale synthetic dataset containing 396k paired transparent–opaque renderings with aligned depth, normals, and object masks. Extensive experiments on synthetic and real-world datasets demonstrate the effectiveness of the proposed method.

In summary, our contributions are as follows: (a).We propose a novel framework \modelname{} that first transforms transparent regions into geometry-consistent opaque images, then enabling further depth prediction for transparent objects using monocular depth models. (b). We introduce a plug-and-play front-end adapter with a mask prediction head to transform transparent regions into geometry-consistent opaque representations. To support training, we construct a large-scale high-quality paired transparent–opaque synthetic dataset, \textbf{SeeClear-396k}. (c). We conduct extensive experiments on both synthetic and real-world datasets, demonstrating consistent improvements over baselines and validating the effectiveness and generality of the proposed method.

%%%%%%%%%%%%%%%%%%%%%%%%%%%%%%%%%%%%%%%%%%%%%%%%%%%%%%%%%%%%
% related work
%%%%%%%%%%%%%%%%%%%%%%%%%%%%%%%%%%%%%%%%%%%%%%%%%%%%%%%%%%%%

\section{Related Work}
\subsection{Monocular Depth Estimation}
Monocular depth estimation infers per-pixel scene depth from an image. Early CNN-based methods \cite{eigen2014depthmappredictionsingle, lee2020cnn} demonstrated direct depth regression, and later transformer architectures such as DPT \cite{ranftl2021visiontransformersdenseprediction} and DINO-derived backbones \cite{oquab2024dinov2learningrobustvisual, simeoni2025dinov3} improved long-range aggregation and depth coherence. Recent depth estimation models \cite{ranftl2020towards, birkl2023midas, keetha2025mapanything}, including ZoeDepth \cite{bhat2023zoedepthzeroshottransfercombining}, Marigold \cite{ke2024repurposingdiffusionbasedimagegenerators}, and Depth Anything \cite{yang2024depth, yang2024depthv2, lin2025depth}, further combine relative-depth pretraining with metric-depth supervision, exploit diffusion priors, and scale training with large-scale pseudo-labeling. These advances improve zero-shot transfer across domains, strengthen metric depth prediction when absolute scale is required, and produce finer and more robust depth maps, making monocular depth practical for many real-world applications. Despite these advances, depth estimation for transparent objects remains challenging, as optical effects such as refraction and reflection violate the standard appearance–geometry correspondence and often lead to unreliable depth predictions \cite{ma2025physliquidphysicsinformeddatasetestimating}. To address this, prior work trains depth models specifically for transparent objects \cite{tosi2024diffusion, costanzino2023learning, dkt2025}. However, these approaches require large-scale paired transparent RGB–depth data for supervision, which is difficult to obtain in reality \cite{sajjan2020cleargrasp, fang2022transcg, chen2022clearpose}. Moreover, these training strategies are model-dependent, requiring retraining for each new depth backbone. Instead, we propose a plug-and-play approach that converts transparent appearances into geometry-consistent opaque images, followed by a monocular depth estimator. This reframes transparent depth estimation as an opacification problem, enabling compatibility with existing depth models without retraining.

\subsection{Transparent Object Segmentation}
Image segmentation \cite{kirillov2023segment, ren2024groundedsamassemblingopenworld, xie2021segmenting, zhang2023shuffletrans, ma2025tosq} provides object-level structural priors that are critical for geometry reasoning, as accurate boundaries help constrain depth inference and prevent cross-region ambiguity. For transparent objects in particular, segmentation has attracted increasing attention due to the prevalence of glass and specular materials in real-world environments \cite{sajjan2020cleargrasp}. However, segmenting transparent objects remains challenging due to complex optical effects such as refraction, reflection, and background distortion \cite{Kalra_2020_CVPR}. Trans2Seg \cite{xie2021segmentingtransparentobjectwild} proposes a transformer model to distinguish transparent materials and introduces the Trans10K dataset as a large-scale benchmark that also supports supervised model training. Trans4Trans~\cite{zhang2021trans4transefficienttransformertransparent} extends this direction with a dual-head transformer for transparent object segmentation to support real-world navigation. However, such methods are designed as task-specific segmentation models, relying on dedicated annotations and optimizing mask quality rather than depth adaptation. Instead, our framework uses general-purpose vision-language segmentation to obtain coarse object-level masks that localize regions requiring appearance opacification before depth estimation.

\subsection{Material Editing}
Broad image editing methods provide useful context but are not designed for material-aware appearance conversion. InstructPix2Pix~\cite{brooks2023instructpix2pix} performs instruction-guided image editing, while AnyDoor~\cite{chen2024anydoor} focuses on object-level image customization. Although these methods can alter object appearance, they do not explicitly model material properties or target transparent-to-opaque conversion. In contrast, we focus on material-level editing to convert transparent objects into opaque appearances. To address image-based material editing, early works such as \cite{liu2017material} adopt physically based inverse rendering to estimate intrinsic properties (e.g., albedo, roughness, metallic) and enable edits through PBR re-rendering, offering interpretability and physical consistency. Careaga et al.~\cite{Careaga_2024} use diffusion models to decompose images into albedo and shading, reducing illumination ambiguity without explicit physical modeling. Recent work Materialist~\cite{wang2025materialistphysicallybasedediting} combines learned material prediction with differentiable rendering for physically consistent editing, including transparency manipulation. Alchemist~\cite{sharma2023alchemistparametriccontrolmaterial} instead uses diffusion models for parametric control over reflectance attributes, but lacks explicit physical constraints. Despite their success, these methods mainly target photorealistic editing or controllable material manipulation rather than edits for downstream depth estimation. In contrast, our method uses paired transparent–opaque guidance to learn appearance transformation that preserves surface geometry and boundaries, producing more reliable inputs for depth prediction.

%%%%%%%%%%%%%%%%%%%%%%%%%%%%%%%%%%%%%%%%%%%%%%%%%%%%%%%%%%%%
% method
%%%%%%%%%%%%%%%%%%%%%%%%%%%%%%%%%%%%%%%%%%%%%%%%%%%%%%%%%%%%
\section{Method}
\label{sec:Methodology}

\begin{figure*}[!t]
  \centering
  \includegraphics[width=\linewidth]{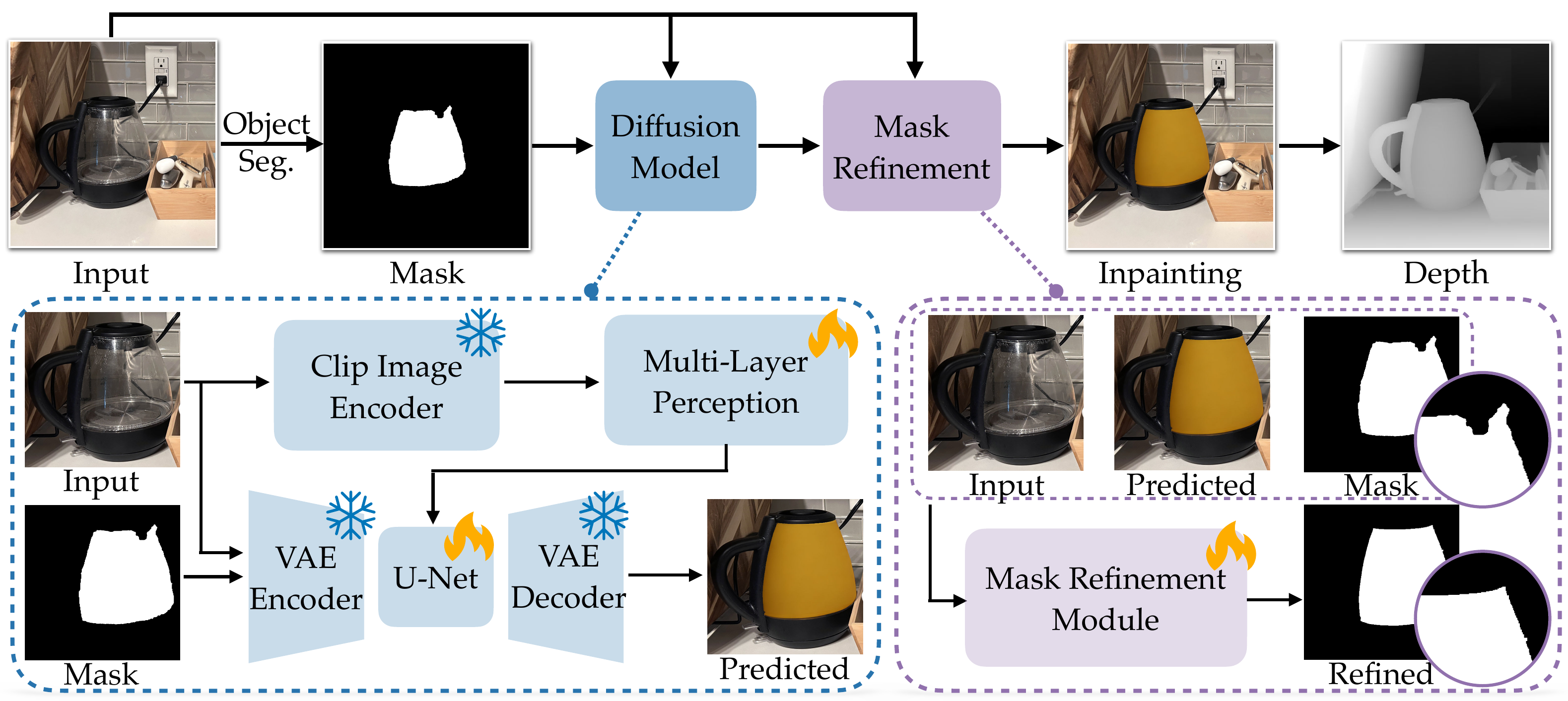}
  \caption{\textbf{Pipeline Overview.} Starting from an image, we first apply a segmentation model to obtain the transparent object mask. Guided by the mask and the image, a latent diffusion model generates an opacified image of the transparent object. A mask refinement module then predicts a soft blending mask to alpha-composite the generated opaque region with the original background, producing the final composited image. The composited image is finally fed into a depth model to estimate accurate depth.}
  \label{fig:pipeline}
\end{figure*}

\begin{figure*}[!t]
    \centering
    \includegraphics[width=\linewidth]{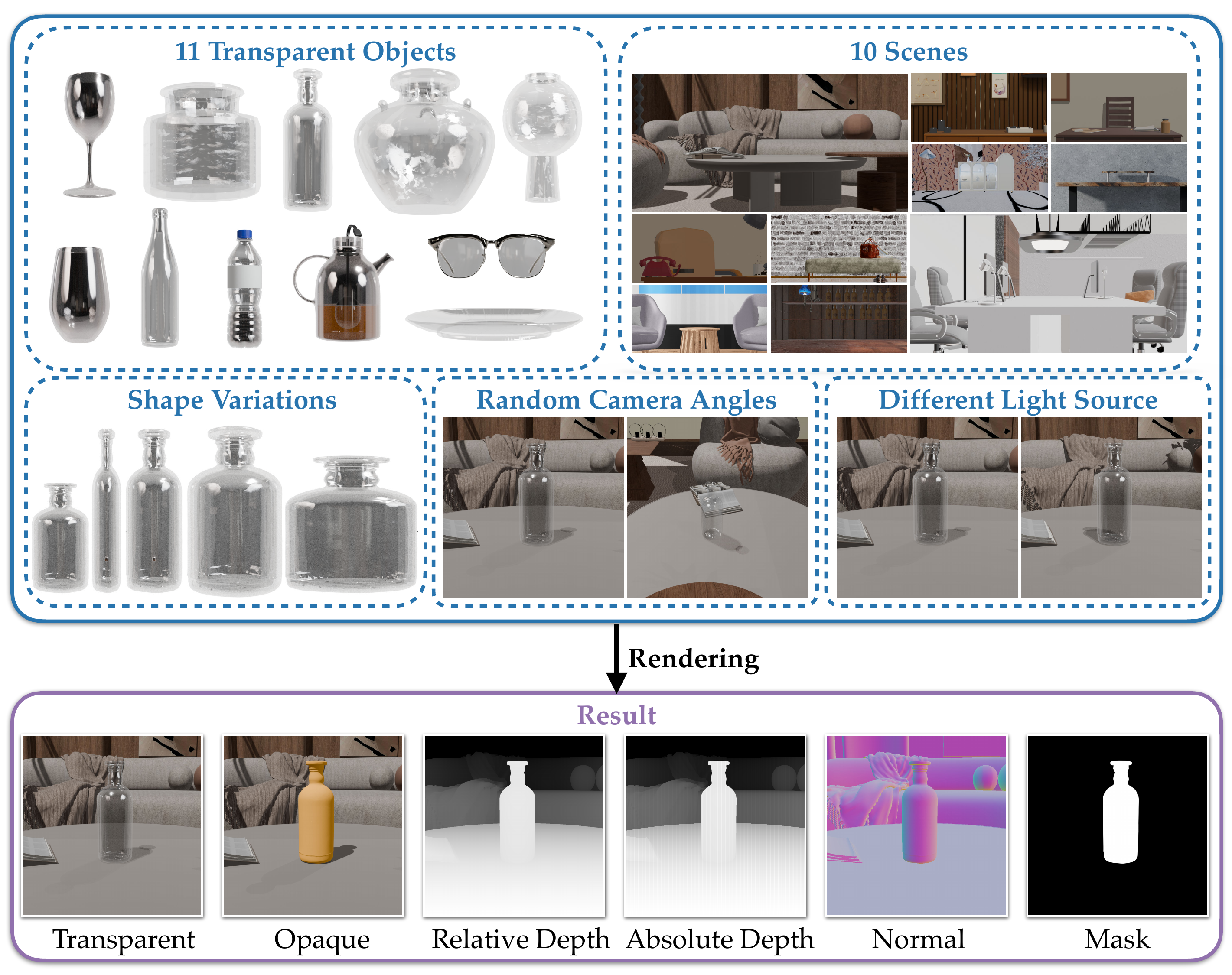}
    \caption{\textbf{Rendering Pipeline.}
    We build the \modelname{}\textbf{-396k} dataset for transparent-object depth estimation.
    For each object–scene configuration, Blender renders paired transparent ($I^{tr}$) and opaque ($I^{op}$) images with identical geometry, camera pose, and illumination. Viewpoint, lighting, and anisotropic shape variations are systematically sampled to produce diverse training data with aligned depth, normals, and masks.}
    \label{fig:transparent_opaque_pairs}
\end{figure*}

Given an input image including transparent objects $I^{tr} \in \mathbb{R}^{H \times W \times 3}$ containing a transparent object, our goal is to estimate its depth map $I^{d} \in \mathbb{R}^{H \times W}$. 
To achieve this, we introduce \modelname{}, a two-stage framework that integrates diffusion-based generative opacification, mask-aware compositing, and off-the-shelf monocular depth estimation. Starting from the input image $I^{tr}$, we first apply transparent-object segmentation models~\cite{zhang2021trans4transefficienttransformertransparent, carion2025sam3segmentconcepts} to obtain the object mask $M^{seg} \in \{0,1\}^{H \times W}$. Conditioned on both $I^{tr}$ and $M^{seg}$, a latent diffusion model transforms refractive appearances into a geometry-consistent opaque image $I^{pred}$. To seamlessly integrate the generated opaque region with the original background, we introduce a mask refinement module that predicts a blending mask $\hat{M}_{refine} \in \{0,1\}^{H \times W}$. The final composited image $I^{blend}$ is obtained via alpha blending between the generated opaque image $I^{pred}$ and the original image $I^{tr}$ using the refined mask $\hat{M}_{refine}$.  The blended image is then fed into an off-the-shelf foundation depth model to produce the final depth map $I^{d}$.

% -----------------------------------------------------------
\subsection{Generative Opacification}
\label{sec:opacification}
% -----------------------------------------------------------
Starting from an input image $I^{tr}$ containing transparent objects, we apply SAM 3~\cite{carion2025sam3segmentconcepts} using visual prompts derived from Trans4Trans~\cite{zhang2021trans4transefficienttransformertransparent} to obtain the transparent object mask $M^{seg}$. Given the input image $I^{tr}$  and transparent object mask $M^{seg}$, this stage generates the opaque object image using a diffusion model. 

\textbf{Conditional Latent Diffusion Formulation.}
We adopt a conditional latent diffusion model to learn the mapping from transparent observations to their opaque counterparts. The ground-truth opaque image $I^{op}$ is first encoded into the latent space of a pre-trained Variational Autoencoder (VAE)~\cite{rombach2022high}, yielding $z_0 = \mathcal{E}(I^{op})$, where $\mathcal{E}$ denotes the VAE encoder. During the forward diffusion process, Gaussian noise is progressively added to $z_0$ to obtain a noisy latent $z_t$. To condition generation on the transparent observation, the denoising network $\epsilon_\theta$ takes as input a concatenated latent tensor
\begin{equation}
  z_{input} = \mathrm{Concat}\!\left(z_t,\; \mathcal{E}(I^{tr}),\; \mathcal{D}_{down}(M^{seg})\right),
\end{equation}

where $I^{tr}$ is the input image containing transparent objects and $M^{seg}$ denotes the transparent object mask. This results in a concatenated feature consisting of the noisy latent $z_t$, the encoded transparent image $\mathcal{E}(I^{tr})$, and a downsampled mask $\mathcal{D}_{down}(M^{seg})$. 
The mask provides explicit spatial guidance, encouraging the model to preserve object geometry while removing refractive distortions. At inference, we adopt UniPC~\cite{zhao2023unipc}, a unified predictor–corrector solver for diffusion ODEs, enabling high-quality sampling in as few as 10 steps without additional network evaluations to generate the opacified image $I^{pred}$.

\textbf{CLIP-based Semantic Conditioning.}
To guide the generation toward an opaque result that preserves the original object shape, an image-level condition encoding object identity is required. Explicit geometric conditions, such as depth or normal maps, would be desirable, but they are unreliable for transparent objects. We therefore use the original transparent-object image $I^{tr}$ as the condition source. Since a transparent object and its diffuse counterpart typically belong to the same semantic category even when their appearances differ significantly, we adopt a CLIP \cite{radford2021learning} image encoder, whose image-text contrastive training yields semantically oriented representations. We retain only the class token as the condition $c$, since patch tokens may encode local appearance details contaminated by background leakage. The class token is then fed into a lightweight MLP to project it into the cross-attention conditioning space of the denoising network $\epsilon_\theta$, yielding the final semantic condition $c$.

\textbf{Training Objectives.}
Following the standard DDPM objective~\cite{ho2020denoising}, we train the 
denoising network $\epsilon_\theta$ to predict the injected noise $\epsilon$ 
at timestep $t$:
\begin{equation}
  \mathcal{L}_{\text{LDM}} = \mathbb{E}_{t, z_0, \epsilon}\left\| 
  \epsilon - \epsilon_\theta\!\left(z_t,\, c,\, t\right) \right\|_2^2.
\end{equation}
While $\mathcal{L}_{\text{LDM}}$ supervises the denoising process in the 
compressed latent space, it does not directly constrain the perceptual quality 
of the decoded output: the VAE's lossy compression discards high-frequency 
details, and element-wise $\ell_2$ supervision is insensitive to perceptually 
significant structures such as sharp edges and fine-grained 
textures~\cite{zhang2018unreasonable}. We therefore introduce a complementary 
perceptual loss computed directly in pixel space using multi-level VGG 
features~\cite{zhang2018unreasonable}:
\begin{equation}
  \mathcal{L}_{\text{LPIPS}} =
  \mathrm{LPIPS}\!\left(\mathcal{D}(z_{pred}) \odot M^{\text{seg}},\;
  I^{op} \odot M^{\text{seg}}\right),
\end{equation}
where $z_{pred}$ denotes the denoised latent estimated from $z_t$ via $\epsilon_\theta$, $\mathcal{D}(z_{pred})$ is the corresponding decoded prediction in pixel space,
$\mathcal{D}(\cdot)$ denotes the VAE decoder, $\odot$ denotes element-wise multiplication, and $M^{\text{seg}}$ confines 
the perceptual supervision to the transparent object region, focusing texture 
fidelity on the opacified content. Following~\cite{ma2026pixelgen}, 
$\mathcal{L}_{\text{LPIPS}}$ is applied only at low-noise timesteps, 
since at high-noise timesteps the decoded prediction deviates significantly 
from the clean image, and perceptual supervision risks reducing sample 
diversity~\cite{ma2026pixelgen}. The overall training objective is $\mathcal{L} = \mathcal{L}_{\text{LDM}} + \lambda\,\mathcal{L}_{\text{LPIPS}}$, where $\lambda$ is a weighting coefficient.

\textbf{Mask augmentation.}\label{par:mask_aug}
Since the segmentation mask $M^{seg}$ at inference may contain boundary inaccuracies, the training mask is randomly perturbed by adding or subtracting small geometric shapes (e.g., circles and rectangles) near object boundaries. This augmentation improves robustness to imperfect masks from the automatic segmentation stage.

% -----------------------------------------------------------
\subsection{Mask Refinement Module}
\label{sec:blending}
% -----------------------------------------------------------
Directly using the decoded image $\mathcal{D}(z_{pred})$ for depth estimation is suboptimal, since VAE encoding-decoding introduces blurring and color drift in non-object regions. To preserve the original scene context, the generated object region should be composited back into the transparent-object image $I^{tr}$. However, latent-space inpainting can introduce boundary halos and color shifts due to VAE nonlinearity~\cite{bradbury2025latentmaskwrongpixelequivalent}, while hard pasting in pixel space with $M^{seg}$ often produces seam artifacts due to boundary inaccuracies in the segmentation mask. We therefore introduce a lightweight Mask Refinement Module (MRM) that outputs a soft blending mask $M_{refine} \in [0,1]^{H \times W}$ for training supervision, which is further binarized at inference to obtain $\hat{M}_{refine} \in \{0,1\}^{H \times W}$ for pixel-level compositing of the generated region with the original image.
The MRM is implemented as a lightweight fully convolutional network consisting of three $3{\times}3$ convolutional layers followed by a $1{\times}1$ projection layer, with GroupNorm and SiLU activations. During training, the module takes as input the concatenation of the opaque-object image $I^{op}$, the transparent-object image $I^{tr}$, and an augmented version of the ground-truth object mask using the same mask augmentation scheme described in Sec.~\ref{sec:opacification}. The module is supervised with a binary cross-entropy loss together with an additional mid-value penalty:
\begin{equation}
  \mathcal{L}_{refine} = \mathcal{L}_{BCE} + \lambda \sum {M}_{refine}(1 - {M}_{refine})
  \label{eq:refine}
\end{equation} 
where $\lambda$ is a weighting coefficient and the second term penalizes mid-range mask values, encouraging confident separation between object and background regions. 
At inference, the opaque-object image and augmented training mask are replaced by the diffusion-generated image $I^{pred}$ and the segmentation mask $M^{seg}$, respectively. The final composited image is then obtained by alpha blending:
\begin{equation}
  I^{blend} = \hat{M}_{refine} \odot I^{pred} + \left(1 - \hat{M}_{refine}\right) \odot I^{tr}.
  \label{eq:blend}
\end{equation}

\subsection{Data Curation}
\label{sec:data}

Our training dataset is designed to improve generalization to in-the-wild transparent objects. % by explicitly covering scenarios where monocular depth estimation fails on transparent materials. 
Following recent transparent-depth benchmarks, we adopt a \emph{failure-mode-driven} construction strategy inspired by Booster~\cite{ramirez2024boosterbenchmarkdepthimages}, which organizes dataset coverage around settings where depth prediction breaks down. We curate objects and scenes to capture three key challenges: non-Lambertian cue distortion (e.g., refraction and textureless glass), depth-definition ambiguity (e.g., object surfaces versus background visible through glass), and complex scene structure (e.g., thin boundaries, multi-layer layouts, and occlusions between transparent and opaque objects). To provide explicit supervision for generative opacification, we render paired transparent and opaque images for each sampled configuration that share identical geometry, camera pose, and illumination (Fig.~\ref{fig:transparent_opaque_pairs}). The transparent image $I^{tr}$ uses a glass-like material model, while the corresponding opaque image $I^{op}$ replaces the transparent shader with a diffuse Lambertian material of fixed color (e.g., \texttt{\#E7A034FF}), isolating appearance changes caused by transparency while keeping all geometric factors fixed. 

To ensure realistic diversity, we curate $11$ everyday transparent objects and furniture placed in $10$ indoor scenes, forming $110$ object–scene units. Each unit is expanded through deterministic sampling of viewpoint, illumination, and shape variation. We sample $10$ camera viewpoints within a predefined angular range, use two lighting setups (canonical overhead and randomized placement), and apply anisotropic scaling under six deformation modes $\{x, y, z, xy, yz, xz\}$ with five magnitudes to diversify object shapes. All parameters are sampled with a fixed random seed for reproducibility. In total, the dataset contains $110$ units with $600$ image sets each, including transparent images $I^{tr}$, opaque images $I^{op}$, absolute depth (OpenEXR), relative depth (PNG), surface normals (PNG), and object masks (PNG), yielding $396{,}000$ rendered images.
%%%%%%%%%%%%%%%%%%%%%%%%%%%%%%%%%%%%%%%%%%%%%%%%%%%%%%%%%%%%
% experiments
%%%%%%%%%%%%%%%%%%%%%%%%%%%%%%%%%%%%%%%%%%%%%%%%%%%%%%%%%%%%
\section{Experimental Evaluation}
\label{sec:exp}
\setcounter{topnumber}{4}
\setcounter{dbltopnumber}{4}
\renewcommand{\dbltopfraction}{0.95}
\renewcommand{\textfraction}{0.05}
\renewcommand{\dblfloatpagefraction}{0.85}

\begin{figure*}[!htbp]
    \centering
    \includegraphics[width=0.79\linewidth]{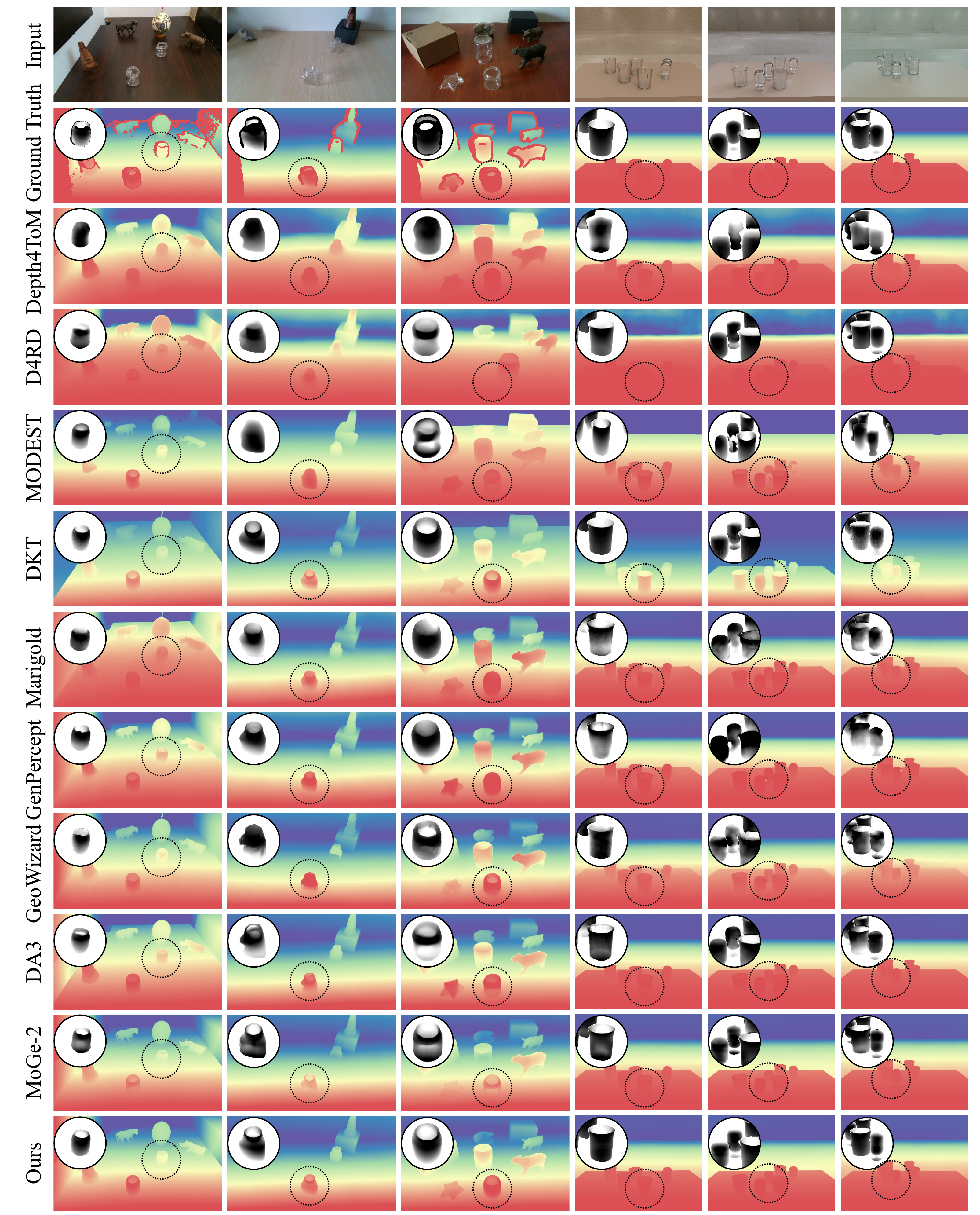}
    \caption{\textbf{Qualitative Comparison.} We evaluate transparent-object depth estimation on  ClearGrasp~\cite{sajjan2020clear} (Columns 1–3) and TransPhy3D~\cite{xu2025diffusion} datasets (Columns 4-6). Compared with the baseline. \modelname{} produces accurate transparent-object depth. Depth maps in the circle are normalized in grayscale.}
    \label{fig:comparison}
\end{figure*}

\begin{table*}[t]
\centering
\caption{\textbf{ClearGrasp Real.} Comparison on the real-world test set across three evaluation regions. 
\textbf{ToM}: transparent-object region. \textbf{All}: full image. \textbf{Other}: non-transparent region. 
$\dagger$: fine-tuned on this dataset. \colorbox{Palette5}{\phantom{x}}: best result.}
\label{tab:cleargrasp}

\resizebox{\linewidth}{!}{%
\renewcommand{\arraystretch}{1.0}
\setlength{\tabcolsep}{4pt}
\setlength{\aboverulesep}{0pt}
\setlength{\belowrulesep}{0pt}
\setlength{\extrarowheight}{0.4pt}

\begin{tabular}{llcccccccc}
\toprule
\textbf{Region} & \textbf{Method} & AbsRel$\downarrow$ & SiLog$\downarrow$ & RMSE$\downarrow$ & iRMSE$\downarrow$ & MAE$\downarrow$ & $\delta_{1.025}$$\uparrow$ & $\delta_{1.05}$$\uparrow$ & $\delta_{1.10}$$\uparrow$ \\
\midrule

\multirow{11}{*}{\textbf{ToM}} 
& Depth4ToM~\cite{costanzino2023learning} & 0.040 & 0.034 & 24.83 & 0.009 & 20.94 & 41.41 & 70.48 & 92.56 \\
& MODEST~\cite{liu2025monocular} & 0.062 & 0.041 & 39.31 & 0.013 & 33.09 & 29.22 & 54.01 & 79.60 \\
& D4RD~\cite{tosi2024diffusion} & 0.069 & 0.031 & 39.95 & 0.014 & 36.18 & 16.21 & 42.70 & 78.25 \\
& DKT$^\dagger$~\cite{xu2025diffusion} & 0.035 & 0.024 & 22.43 & 0.008 & 18.59 & \cellcolor{Palette5}\textbf{52.69} & \cellcolor{Palette5}\textbf{81.29} & 94.04 \\
& Marigold~\cite{ke2025marigold} & 0.047 & 0.036 & 28.10 & 0.011 & 24.31 & 38.58 & 64.49 & 88.10 \\
& GenPercept~\cite{xu2024matters} & 0.039 & 0.028 & 24.09 & 0.009 & 20.30 & 47.79 & 75.21 & 92.71 \\
& GeoWizard~\cite{fu2024geowizard} & 0.052 & 0.029 & 31.54 & 0.012 & 27.54 & 34.86 & 62.25 & 87.87 \\
& DA3~\cite{lin2025depth} & 0.038 & 0.028 & 24.03 & 0.008 & 19.95 & 46.89 & 75.82 & 92.24 \\
& MoGe-2~\cite{wang2025moge} & 0.101 & 0.035 & 56.83 & 0.019 & 52.96 & 5.63 & 17.81 & 60.01 \\

\midrule
\rowcolor{OursColor} 
& Ours (DA3) & \cellcolor{Palette5}\textbf{0.033} & \cellcolor{Palette5}\textbf{0.023} & \cellcolor{Palette5}\textbf{21.45} & \cellcolor{Palette5}\textbf{0.007} & \cellcolor{Palette5}\textbf{17.94} & 51.27 & 78.13 & \cellcolor{Palette5}\textbf{95.63} \\

\rowcolor{OursColor} 
& Ours (MoGe-2) & 0.043 & 0.024 & 26.62 & 0.009 & 22.94 & 42.75 & 71.43 & 91.55 \\

\midrule

\multirow{11}{*}{\textbf{All}} 
& Depth4ToM~\cite{costanzino2023learning} & 0.044 & 0.055 & 44.94 & 0.009 & 29.74 & 51.14 & 71.87 & 85.68 \\
& MODEST~\cite{liu2025monocular} & 0.087 & 0.123 & 111.43 & 0.019 & 58.05 & 59.23 & 72.61 & 79.44 \\
& D4RD~\cite{tosi2024diffusion} & 0.026 & 0.037 & 27.71 & 0.007 & 16.65 & 74.08 & 86.20 & 93.27 \\
& DKT$^\dagger$~\cite{xu2025diffusion} & 0.111 & 0.144 & 137.48 & 0.021 & 70.21 & 61.05 & 74.63 & 81.00 \\
& Marigold~\cite{ke2025marigold} & 0.049 & 0.062 & 48.64 & 0.011 & 31.76 & 49.05 & 74.36 & 87.07 \\
& GenPercept~\cite{xu2024matters} & 0.034 & 0.042 & 32.89 & 0.008 & 22.10 & 59.75 & 83.34 & 92.58 \\
& GeoWizard~\cite{fu2024geowizard} & 0.027 & 0.036 & 26.80 & 0.008 & 17.33 & 75.65 & 86.41 & 93.93 \\
& DA3~\cite{lin2025depth} & 0.027 & 0.046 & 38.88 & 0.007 & 18.19 & 79.53 & 88.96 & 96.21 \\
& MoGe-2~\cite{wang2025moge} & 0.020 & 0.031 & 22.30 & 0.006 & 13.31 & 77.51 & 89.31 & 96.82 \\

\midrule
\rowcolor{OursColor} 
& Ours (DA3) & 0.025 & 0.041 & 34.85 & 0.006 & 16.87 & 79.93 & 89.28 & 96.58 \\

\rowcolor{OursColor} 
& Ours (MoGe-2) & \cellcolor{Palette5}\textbf{0.016} & \cellcolor{Palette5}\textbf{0.022} & \cellcolor{Palette5}\textbf{17.87} & \cellcolor{Palette5}\textbf{0.004} & \cellcolor{Palette5}\textbf{11.19} & \cellcolor{Palette5}\textbf{80.63} & \cellcolor{Palette5}\textbf{92.77} & \cellcolor{Palette5}\textbf{99.05} \\

\midrule

\multirow{11}{*}{\textbf{Other}} 
& Depth4ToM~\cite{costanzino2023learning} & 0.044 & 0.055 & 45.56 & 0.009 & 30.09 & 51.74 & 72.23 & 85.55 \\
& MODEST~\cite{liu2025monocular} & 0.087 & 0.124 & 113.28 & 0.019 & 59.09 & 60.50 & 73.52 & 79.56 \\
& D4RD~\cite{tosi2024diffusion} & 0.023 & 0.033 & 25.71 & 0.006 & 15.54 & 76.76 & 88.53 & 94.44 \\
& DKT$^\dagger$~\cite{xu2025diffusion} & 0.113 & 0.146 & 139.87 & 0.021 & 72.04 & 61.39 & 74.35 & 80.49 \\
& Marigold~\cite{ke2025marigold} & 0.049 & 0.062 & 49.09 & 0.011 & 32.04 & 49.60 & 74.97 & 87.25 \\
& GenPercept~\cite{xu2024matters} & 0.034 & 0.042 & 33.07 & 0.008 & 22.17 & 60.27 & 83.79 & 92.69 \\
& GeoWizard~\cite{fu2024geowizard} & 0.025 & 0.034 & 26.28 & 0.007 & 16.91 & 77.40 & 87.45 & 94.27 \\
& DA3~\cite{lin2025depth} & 0.026 & 0.045 & 38.78 & 0.006 & 18.04 & 81.19 & 89.94 & 96.58 \\
& MoGe-2~\cite{wang2025moge} & 0.016 & 0.021 & 17.97 & \cellcolor{Palette5}\textbf{0.003} & 11.24 & 80.61 & 92.55 & 99.03 \\

\midrule
\rowcolor{OursColor} 
& Ours (DA3) & 0.024 & 0.041 & 34.93 & 0.006 & 16.87 & 81.07 & 89.73 & 96.71 \\

\rowcolor{OursColor} 
& Ours (MoGe-2) & \cellcolor{Palette5}\textbf{0.015} & \cellcolor{Palette5}\textbf{0.020} & \cellcolor{Palette5}\textbf{17.12} & \cellcolor{Palette5}\textbf{0.003} & \cellcolor{Palette5}\textbf{10.71} & \cellcolor{Palette5}\textbf{82.17} & \cellcolor{Palette5}\textbf{93.64} & \cellcolor{Palette5}\textbf{99.44} \\

\bottomrule
\end{tabular}
}
\end{table*}

\begin{table*}[t]
\centering
\begin{subtable}[t]{0.49\linewidth}
\centering
\caption{\textbf{TransPhy3D (synthetic).} Comparison on the synthetic test set. 
\colorbox{Palette5}{\phantom{x}}: best result. $\dagger$: fine-tuned on this dataset.}
\label{tab:transphy3d}
\makebox[\linewidth][c]{\resizebox{1.1\linewidth}{!}{%
\renewcommand{\arraystretch}{1.0}
\setlength{\tabcolsep}{2.5pt}
\setlength{\aboverulesep}{0pt}
\setlength{\belowrulesep}{0pt}
\setlength{\extrarowheight}{0.4pt}
\begin{tabular}{>{\raggedright\arraybackslash}p{2.3cm} *{8}{>{\centering\arraybackslash}p{1.05cm}}}
\toprule
\textbf{Method} & AbsRel$\downarrow$ & SiLog$\downarrow$ & RMSE$\downarrow$ & iRMSE$\downarrow$ & MAE$\downarrow$ & $\delta_{1.025}$$\uparrow$ & $\delta_{1.05}$$\uparrow$ & $\delta_{1.10}$$\uparrow$ \\
\midrule
Depth4ToM~\cite{costanzino2023learning} & 0.027 & 0.051 & 60.13 & 0.006 & 30.27 & 72.39 & 88.79 & 96.29 \\
MODEST~\cite{liu2025monocular} & 0.035 & 0.061 & 67.54 & 0.007 & 37.40 & 66.97 & 85.26 & 93.81 \\
D4RD~\cite{tosi2024diffusion} & 0.017 & 0.039 & 51.76 & 0.004 & 23.67 & 90.94 & 96.40 & 98.20 \\
DKT$^\dagger$~\cite{xu2025diffusion} & 0.014 & 0.029 & \cellcolor{Palette5}\textbf{33.20} & \cellcolor{Palette5}\textbf{0.003} & \cellcolor{Palette5}\textbf{13.88} & 89.20 & 96.33 & 98.57 \\
Marigold~\cite{ke2025marigold} & 0.027 & 0.044 & 50.72 & 0.005 & 27.88 & 66.37 & 88.76 & 97.36 \\
GenPercept~\cite{xu2024matters} & 0.037 & 0.059 & 58.19 & 0.008 & 33.31 & 53.56 & 78.34 & 92.78 \\
GeoWizard~\cite{fu2024geowizard} & 0.025 & 0.045 & 55.88 & 0.005 & 29.63 & 74.64 & 92.74 & 97.91 \\
DA3~\cite{lin2025depth} & 0.023 & 0.043 & 53.25 & 0.005 & 25.27 & 76.00 & 93.91 & 97.99 \\
MoGe-2~\cite{wang2025moge} & 0.016 & 0.030 & 39.78 & \cellcolor{Palette5}\textbf{0.003} & 19.21 & 89.58 & 96.17 & 98.62 \\
\midrule
\rowcolor{OursColor} Ours (DA3) & 0.017 & 0.037 & 47.46 & 0.004 & 21.11 & 88.01 & 96.42 & 98.29 \\
\rowcolor{OursColor} Ours (MoGe-2) & \cellcolor{Palette5}\textbf{0.012} & \cellcolor{Palette5}\textbf{0.028} & 38.91 & \cellcolor{Palette5}\textbf{0.003} & 16.81 & \cellcolor{Palette5}\textbf{94.43} & \cellcolor{Palette5}\textbf{97.50} & \cellcolor{Palette5}\textbf{98.67} \\
\bottomrule
\end{tabular}
}}
\end{subtable}
\hspace{0.03\linewidth}
\begin{subtable}[t]{0.47\linewidth}
\centering
\caption{\textbf{Ablation study} on ClearGrasp Real (ToM region). 
All variants use Depth Anything V3 as the depth backbone.
\colorbox{Palette5}{\phantom{x}}: best result.}
\label{tab:ablation}
\makebox[\linewidth][c]{\resizebox{1.1\linewidth}{!}{%
\renewcommand{\arraystretch}{1.0}
\setlength{\tabcolsep}{2.5pt}
\setlength{\aboverulesep}{0pt}
\setlength{\belowrulesep}{0pt}
\setlength{\extrarowheight}{0.4pt}
\begin{tabular}{>{\raggedright\arraybackslash}p{2.5cm} *{8}{>{\centering\arraybackslash}p{1.05cm}}}
\toprule
\textbf{Method} & AbsRel$\downarrow$ & SiLog$\downarrow$ & RMSE$\downarrow$ & iRMSE$\downarrow$ & MAE$\downarrow$ & $\delta_{1.025}$$\uparrow$ & $\delta_{1.05}$$\uparrow$ & $\delta_{1.10}$$\uparrow$ \\
\midrule
w/ BBox & 0.035 & 0.025 & 22.96 & 0.008 & 18.99 & 50.90 & 76.05 & 93.77 \\
w/ Point & 0.034 & 0.024 & 22.38 & 0.008 & 18.68 & 49.87 & 77.32 & 94.47 \\
\midrule
w/ DINOv2 CLS & 0.037 & 0.025 & 23.85 & 0.008 & 20.08 & 49.78 & 75.84 & 92.73 \\
w/ DINOv2 full & 0.034 & 0.024 & 22.35 & 0.008 & 18.67 & 51.53 & 78.80 & 94.32 \\
w/ CLIP full & 0.034 & 0.024 & 22.30 & 0.008 & 18.58 & 48.91 & 77.99 & 95.08 \\
\midrule
w/o MRM & \cellcolor{Palette5}\textbf{0.033} & 0.025 & 21.73 & 0.008 & 18.00 & 50.48 & \cellcolor{Palette5}\textbf{80.66} & 94.58 \\
w/o blending & 0.037 & 0.026 & 23.60 & 0.008 & 19.83 & 45.45 & 76.19 & 93.46 \\
\midrule
w/ $\mathcal{L}_{\text{LDM}}$ & 0.034 & 0.024 & 22.30 & 0.007 & 18.61 & 50.46 & 76.87 & 95.35 \\
w/ $\mathcal{L}_{\text{LDM}}$+$\mathcal{L}_{\mathrm{grad}}$ & 0.034 & 0.023 & 22.21 & 0.007 & 18.56 & 50.00 & 77.22 & 95.46 \\
w/ $\mathcal{L}_{\text{LDM}}$+$\mathcal{L}_{\mathrm{L1}}$ & 0.034 & 0.023 & 21.81 & 0.007 & 18.19 & 50.10 & \cellcolor{Palette5}\textbf{79.05} & 95.16 \\
\midrule
DA3 (baseline) & 0.038 & 0.028 & 24.03 & 0.008 & 19.95 & 46.89 & 75.82 & 92.24 \\
Solid-color & 0.053 & 0.035 & 33.09 & 0.012 & 28.11 & 34.00 & 61.52 & 84.64 \\
\midrule
\rowcolor{OursColor}\textbf{Ours} & \cellcolor{Palette5}\textbf{0.033} & \cellcolor{Palette5}\textbf{0.023} & \cellcolor{Palette5}\textbf{21.45} & \cellcolor{Palette5}\textbf{0.007} & \cellcolor{Palette5}\textbf{17.94} & \cellcolor{Palette5}\textbf{51.27} & 78.13 & \cellcolor{Palette5}\textbf{95.63} \\
\bottomrule
\end{tabular}
}}
\end{subtable}
\end{table*}

\begin{figure*}[htbp]
    \centering
    \includegraphics[width=0.79\linewidth]{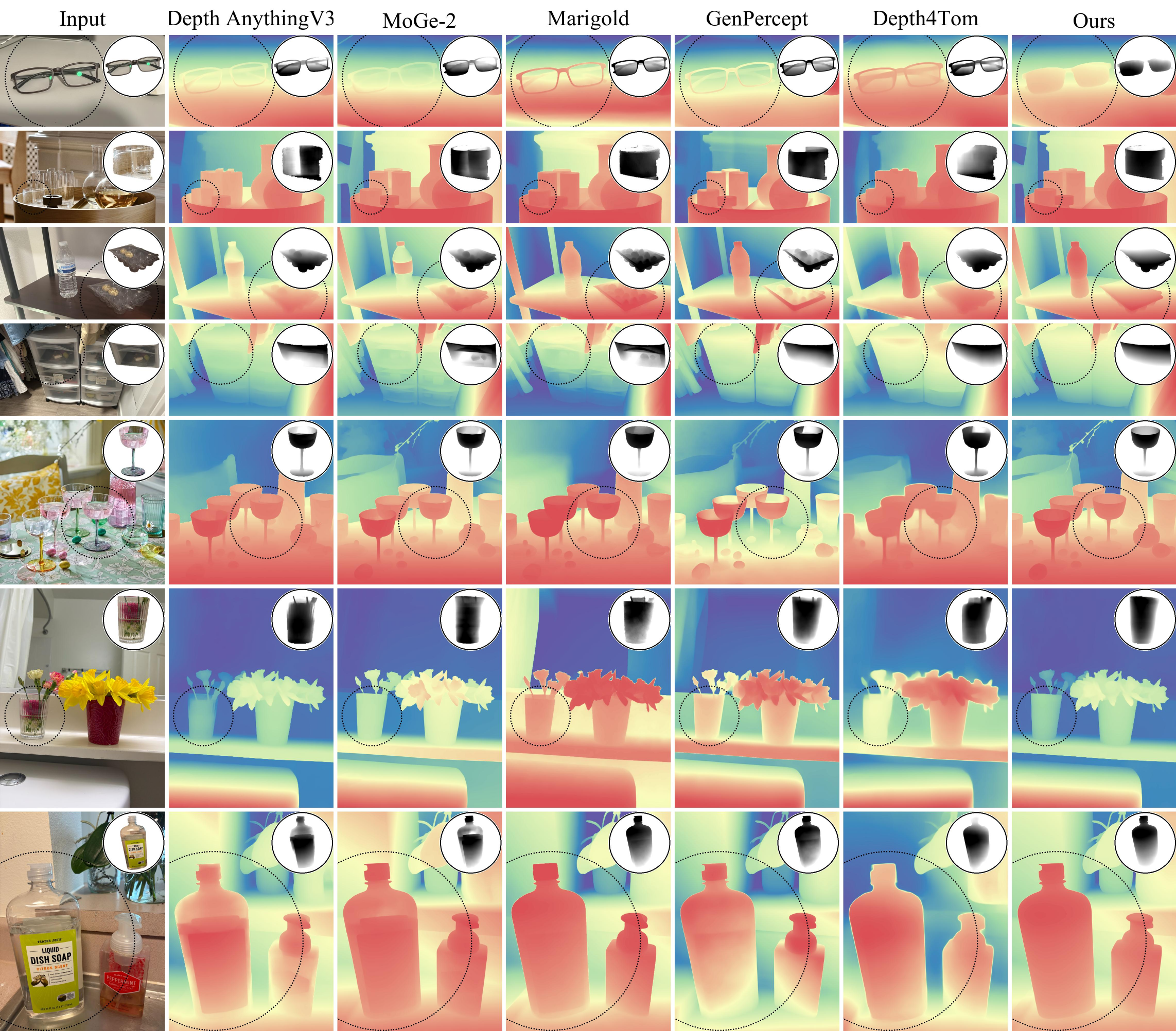}
    \caption{\textbf{In-the-Wild Qualitative Comparison (Part~I).} Scenes include a pair of glasses (row~1), multiple transparent objects with liquid and inter-object occlusion (row~2), a transparent container with an out-of-distribution shape enclosing opaque chocolate objects under a plastic film (row~3), a transparent plastic drawer cabinet containing opaque objects (row~4), multiple transparent objects with inter-object occlusion and liquid contents (row~5), a single transparent glass containing liquid and plants extending from inside to outside (row~6), and multiple transparent objects with opaque labels and liquid contents (row~7). Depth4ToM produces predictions with blurrier boundaries compared with other methods; in rows~1, 5, and~6 the transparent surfaces appear translucent, and in rows~2 and~7 the depth fails to reflect object surface geometry such as cup openings and bottle shapes. Depth Anything~V3 and MoGe-2 exhibit pronounced translucency across all rows. Marigold and GenPercept both show translucency in rows~1, 3, 5, 6, and~7; additionally, Marigold exhibits translucency in row~4, and GenPercept exhibits translucency in row~2. \modelname{} produces no translucency across all scenes, accurately recovers object surface geometry, and preserves seamless depth transitions at material boundaries.}
    \label{fig:wild-1}
\end{figure*}

\begin{figure*}[htbp]
    \centering
    \includegraphics[width=0.79\linewidth]{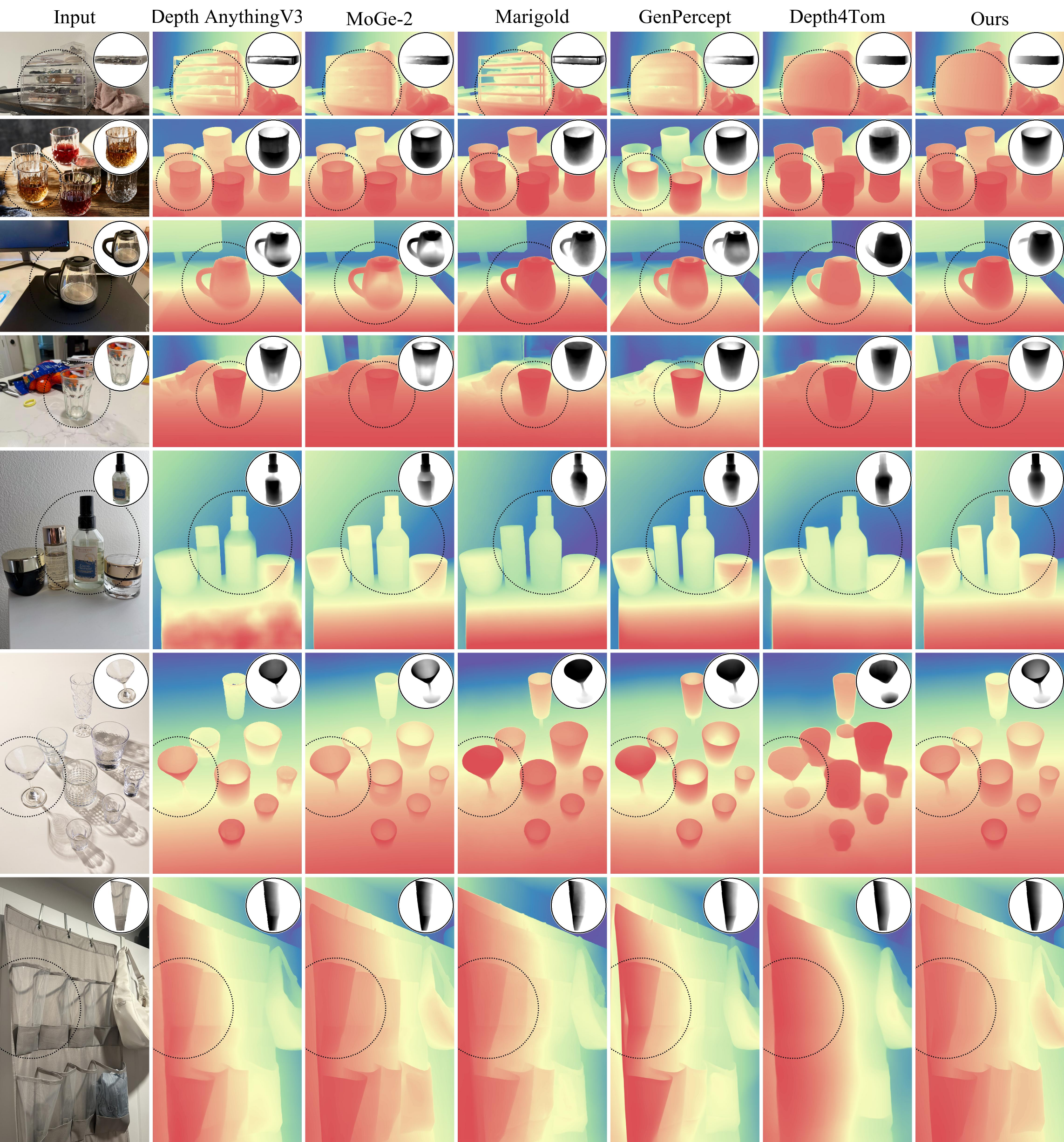}
    \caption{\textbf{In-the-Wild Qualitative Comparison (Part~II).} Scenes include a transparent storage box containing opaque objects (row~1), multiple transparent objects with liquid and inter-object occlusion (row~2), objects with mixed transparent and opaque materials (row~3), a single transparent object (row~4), multiple transparent objects with opaque labels and liquid contents (row~5), multiple transparent objects (row~6), and an out-of-distribution scene combining transparent and opaque materials (row~7). Depth4ToM produces blurry boundaries throughout; in rows~2--6 the depth fails to reflect object surface geometry such as cup openings, kettle shapes, and the surface shape of bottles and containers, and in row~7 transparent objects show no discernible shape in the depth map. Depth Anything~V3 and MoGe-2 exhibit translucency across all rows. Marigold and GenPercept both exhibit translucency in rows~1, 3, and~7; additionally, GenPercept shows translucency in row~4. Marigold produces uneven depth transitions in rows~2, 3, and~5, and GenPercept exhibits similar unevenness in rows~3 and~5. In rows~4 and~6, Marigold produces depth maps that do not match object geometry (e.g., missing cup openings), and GenPercept shows the same issue in row~6. \modelname{} accurately recovers object geometry including openings and surface contours, produces consistent depth under occlusion, and maintains seamless transitions at material boundaries.}
    \label{fig:wild-2}
\end{figure*}

\begin{figure*}[!htbp]
\centering
\begin{minipage}[t]{0.49\linewidth}
    \centering
    \includegraphics[width=\linewidth, trim=0 0 0 33, clip]{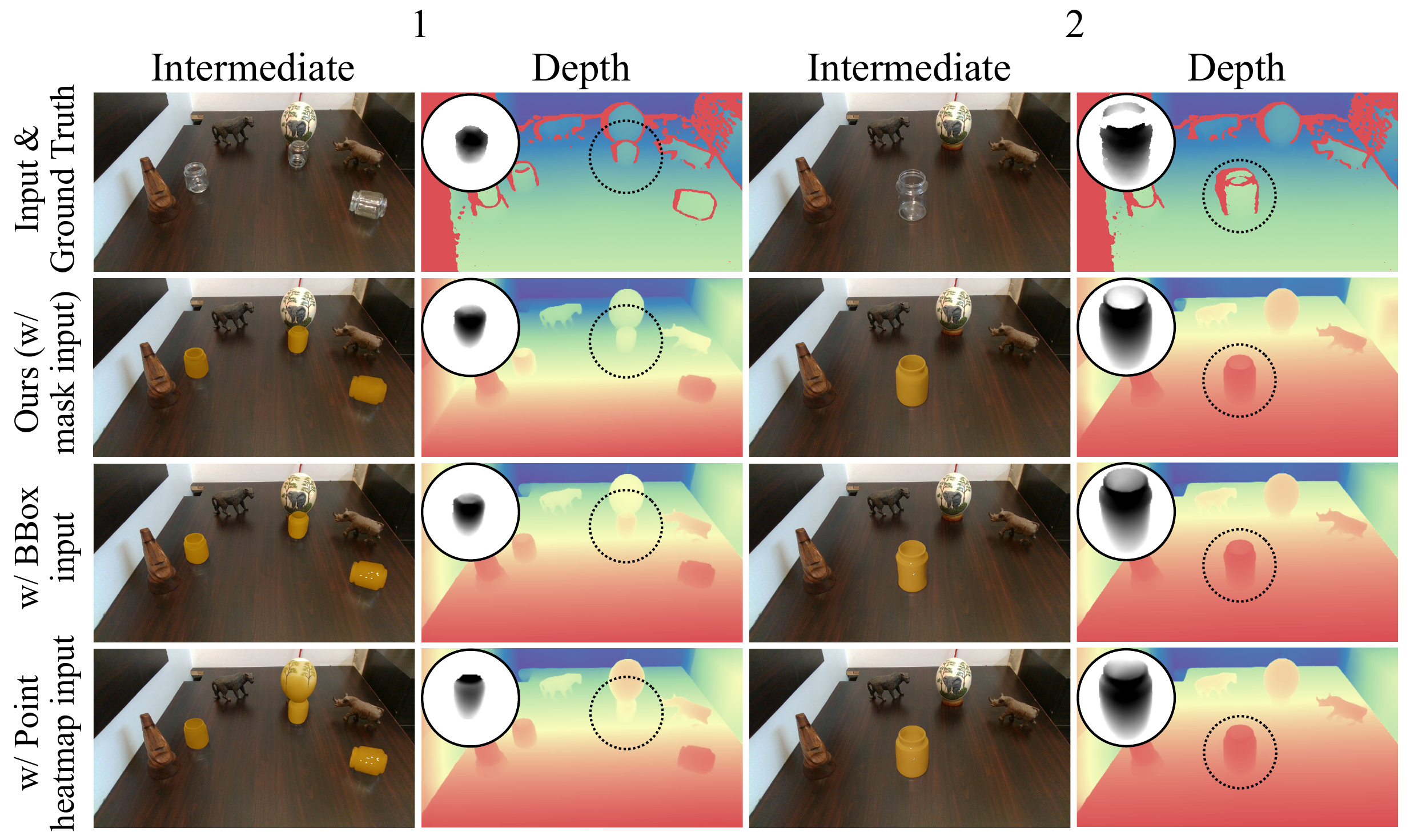}
    \caption{\textbf{Transparent Object Localization Ablation.} The mask $M^{seg}$ provides precise spatial guidance, while bounding boxes and point heatmaps cause inaccurate opaque images and depths. Depth maps in the circle are normalized.}
    \label{fig:ablation2}
\end{minipage}\hspace{0.005\linewidth}%
\begin{minipage}[t]{0.49\linewidth}
    \centering
    \includegraphics[width=\linewidth, trim=0 0 0 32, clip]{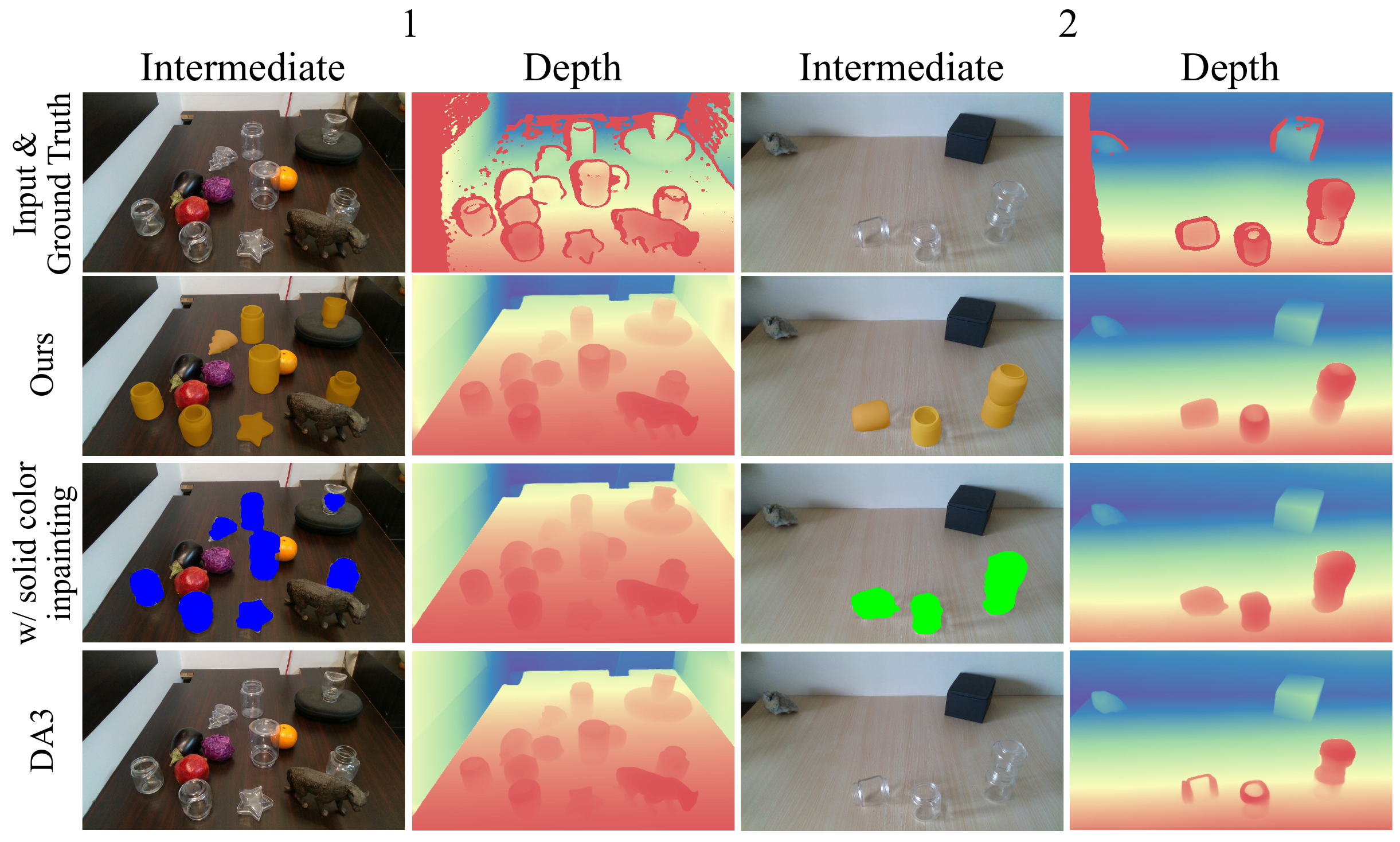}
    \caption{\textbf{Opacified Image Ablation.} Flat-color opaque images lacks shading cues, causing large depth errors, while our generative opacification restores plausible appearance for depth estimation and predicts accurate depth maps.}
    \label{fig:ablation4}
\end{minipage}
\end{figure*}

\begin{figure*}[!htbp]
    \centering
    \begin{minipage}[t]{0.32\linewidth}
        \centering
        \vspace{0pt}
        \includegraphics[width=\linewidth]{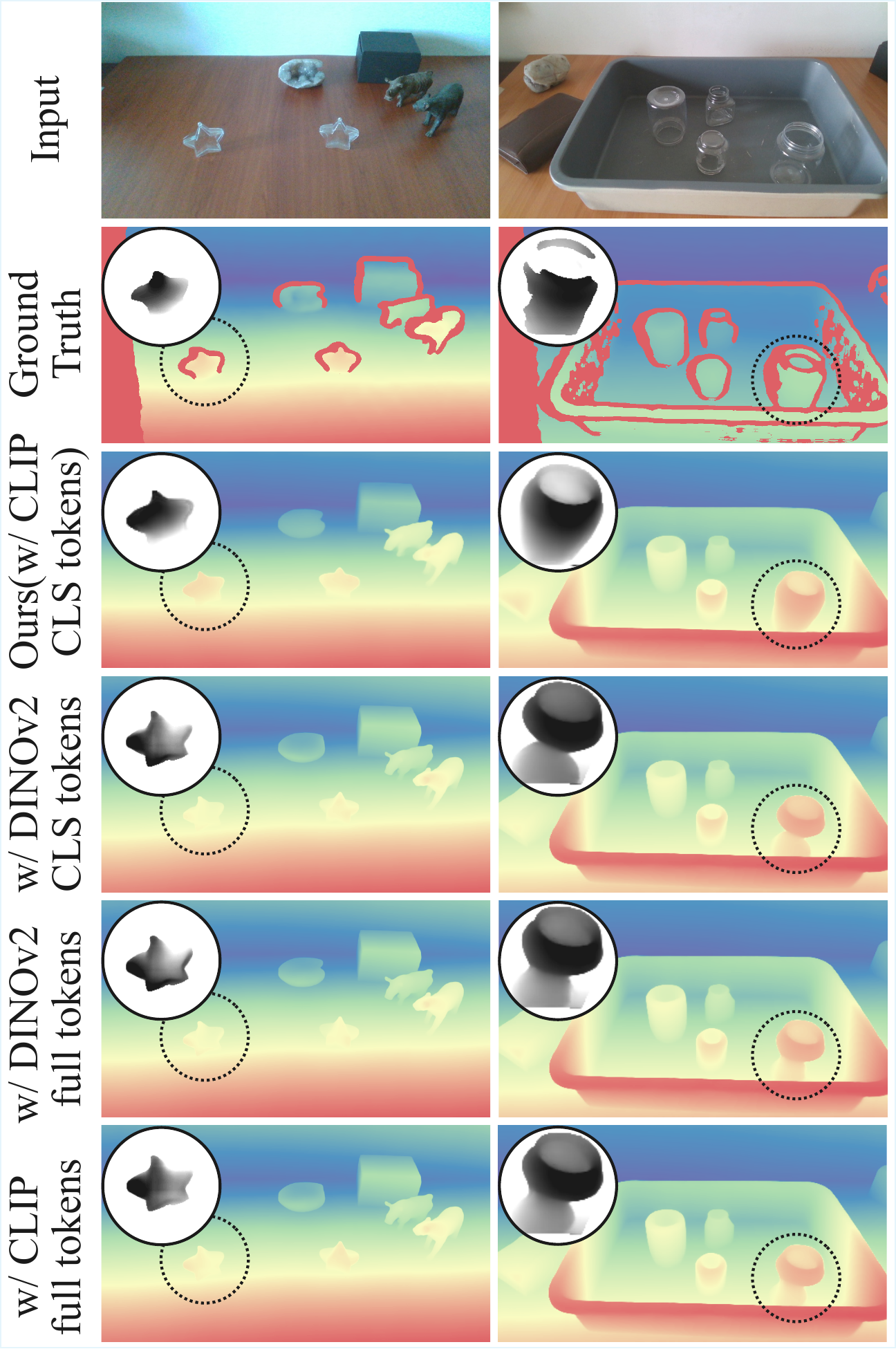}
        \caption{\textbf{Image Encoder Ablation.} Comparison of CLIP and DINOv2 conditioning tokens. The CLIP class token provides the most stable signal and achieves the best depth accuracy.}
        \label{fig:ablation1}
    \end{minipage}\hfill
    \begin{minipage}[t]{0.32\linewidth}
        \centering
        \vspace{0pt}
        \includegraphics[width=\linewidth]{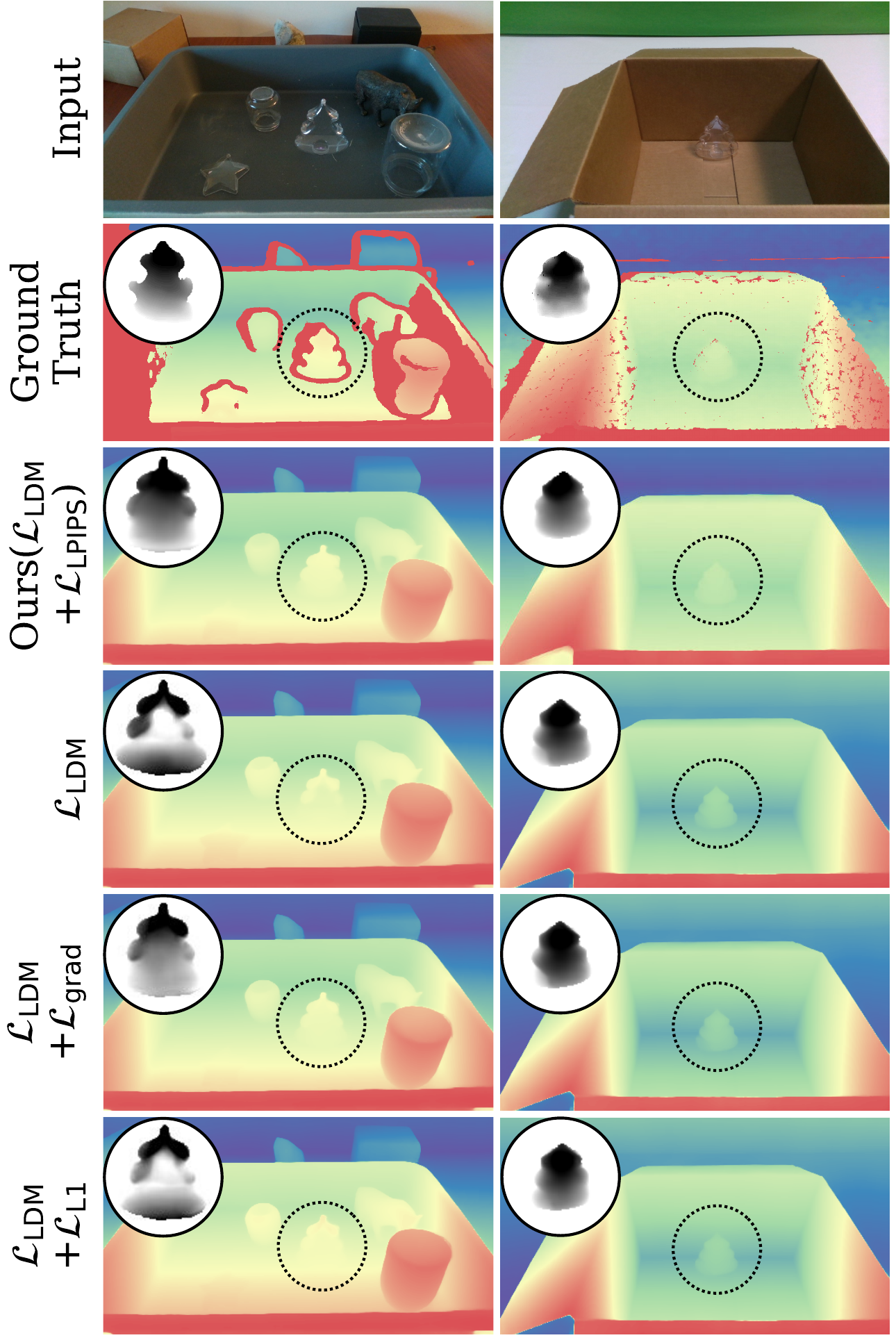}
        \caption{\textbf{Perceptual Loss Ablation.} Without LPIPS supervision, variants produce incomplete opacification with translucent interiors or under-opacified boundaries. Our LPIPS variant produces the most geometry-consistent opaque appearances.}
        \label{fig:ablation-loss}
    \end{minipage}\hfill
    \begin{minipage}[t]{0.34\linewidth}
        \centering
        \vspace{0pt}
        \includegraphics[width=\linewidth]{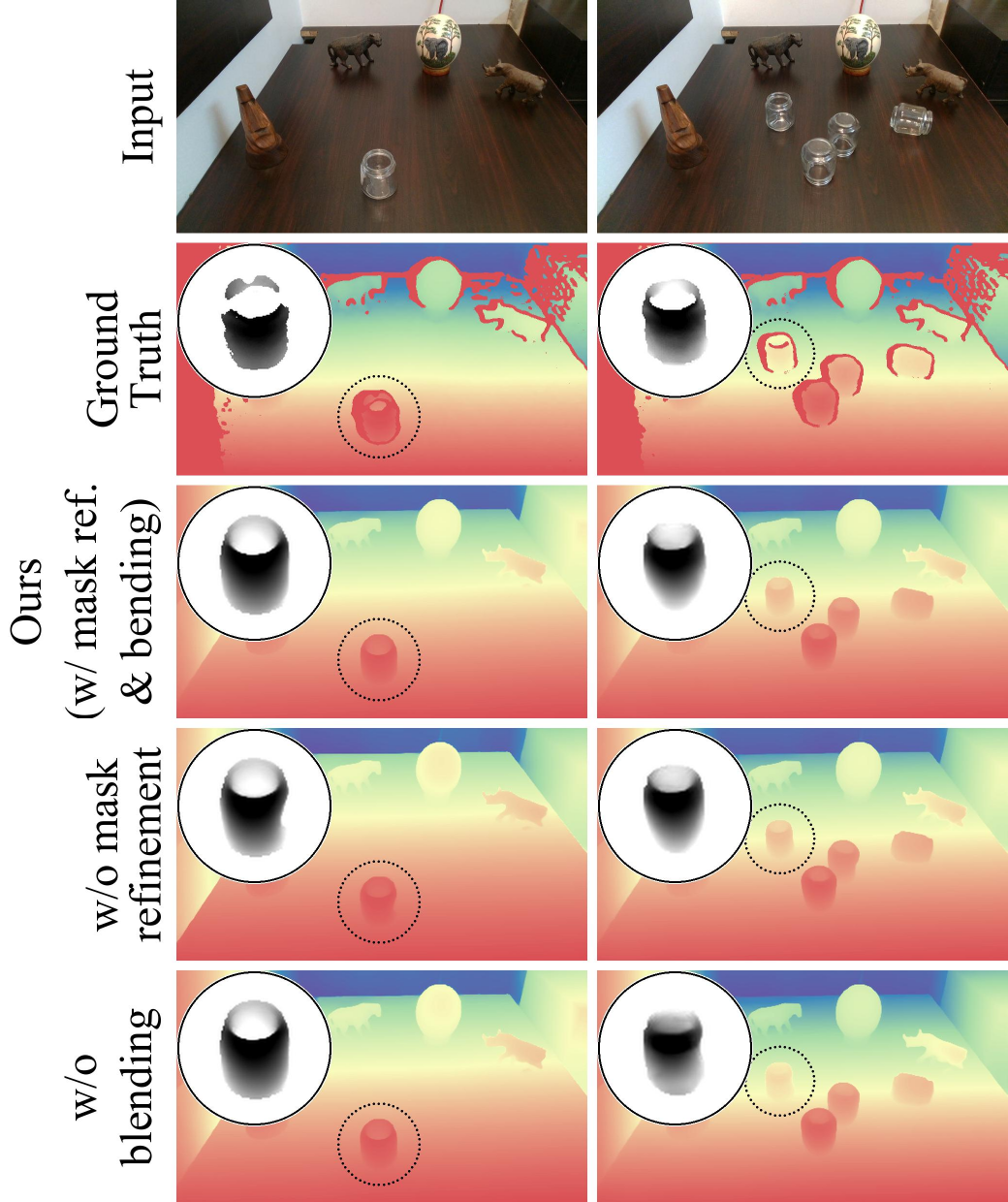}
        \caption{\textbf{Image Compositing Ablation.} Removing MRM causes blurry boundaries and incomplete opacification, while removing alpha blending introduces background drift from the generated image, causing inaccurate depth predictions.}
        \label{fig:ablation3}
    \end{minipage}
\end{figure*}

% ----------------------------------------------------------
\textbf{Implementation Details}
\label{sec:impl}
% ----------------------------------------------------------
The opacification network is implemented as a latent diffusion model~\cite{rombach2022high} initialized from Paint-by-Example~\cite{yang2023paint} and fine-tuned on the \modelname{}-396k dataset described in Sec.~\ref{sec:data}.  For transparent object localization, we adopt a coarse-to-fine segmentation strategy. Trans4Trans~\cite{zhang2021trans4transefficienttransformertransparent} first provides an initial localization prior, which serves as a spatial prompt for SAM~3~\cite{carion2025sam3segmentconcepts} to produce a refined transparent-object mask $M^{seg}$. Training uses a base learning rate of $1\times10^{-5}$ with a linear warm-up schedule and a batch size of 4, and runs for 328,339 iterations on four NVIDIA H100 NVL GPUs. All training images are resized to $512\times512$. At inference, the model is applied per instance: for each transparent object, a local patch centered on its bounding box is cropped from the input image, resized to $512\times512$, processed by the diffusion model, and composited back into the original image using the Mask Refinement Module. The diffusion process is solved using the UniPC sampler~\cite{zhao2023unipc} with 10 steps. For depth estimation, we explore two off-the-shelf monocular depth models, Depth Anything~V3~\cite{lin2025depth} and MoGe-2~\cite{wang2025moge}. Neither backbone is fine-tuned, and all evaluations are conducted in a zero-shot setting.

% ----------------------------------------------------------
\textbf{Baselines}
\label{sec:baselines} We compare \modelname{} with several open-source state-of-the-art methods for transparent-object depth estimation, including Depth4ToM~\cite{costanzino2023learning}, Diffusion4RobustDepth (D4RD)~\cite{tosi2024diffusion}, MODEST~\cite{liu2025monocular}, and DKT~\cite{xu2025diffusion}. In addition, we compare with general-purpose monocular depth estimators. These include diffusion-based depth models such as Marigold~\cite{ke2025marigold}, GenPercept~\cite{xu2024matters}, and GeoWizard~\cite{fu2024geowizard}, which leverage pre-trained generative diffusion priors for depth prediction. We also evaluate discriminative depth models, including Depth Anything~V3 (DA3)~\cite{lin2025depth} and MoGe-2~\cite{wang2025moge}, which are feed-forward ViT-based estimators trained on large-scale depth datasets. Note that DKT is fine-tuned on both evaluation datasets and therefore does not operate in a zero-shot setting. Other methods, including the proposed method, are evaluated without training on the evaluation datasets. To compare with DKT in a zero-shot setting, we also evaluate on additional datasets where DKT is not trained, with results reported in the Appendix.

% ----------------------------------------------------------
\subsection{Quantitative Evaluation}
\label{sec:quant}
% ----------------------------------------------------------
To evaluate transparent-object depth estimation, we evaluate transparent-object depth estimation on both real-world and synthetic benchmarks, including \textit{ClearGrasp}~\cite{sajjan2020clear}, an RGB-D dataset of transparent objects with sensor-captured depth, and \textit{TransPhy3D}~\cite{xu2025diffusion}, a synthetic dataset with physically rendered depth for transparent objects. Following prior work~\cite{xu2024matters}, affine-invariant and disparity-output models are aligned to metric ground-truth depth via least-squares regression before computing evaluation metrics. For disparity models, the regression is performed in the disparity domain and converted back to depth. Evaluation on \textit{ClearGrasp}~\cite{sajjan2020clear} uses three disjoint regions: the transparent-object region (\textbf{ToM}), the full image (\textbf{All}), and the non-transparent region (\textbf{Other}). For \textit{TransPhy3D}~\cite{xu2025diffusion}, evaluation is performed only on the full image, as no ground-truth object masks are provided for region-specific evaluation. We report standard depth-estimation metrics including AbsRel~($\downarrow$), SiLog~($\downarrow$), RMSE~($\downarrow$, mm), iRMSE~($\downarrow$, $\text{mm}^{-1}$), MAE~($\downarrow$, mm), and threshold accuracies $\delta{<}1.025$, $\delta{<}1.05$, and $\delta{<}1.10$~($\uparrow$), measuring relative depth error (AbsRel), scale-invariant logarithmic error (SiLog), root-mean-square error sensitive to large depth deviations (RMSE), inverse-depth error emphasizing near-range accuracy (iRMSE), mean absolute error (MAE), and the percentage of pixels whose predicted depth falls within a multiplicative threshold of the ground truth ($\delta$ metrics).

\textbf{Quantitative Results on ClearGrasp.} As shown in Tab.~\ref{tab:cleargrasp}, \modelname{} outperforms all baselines on the \textbf{All} and \textbf{Other} regions across all metrics on the real-world ClearGrasp dataset, while also surpassing most baselines on the \textbf{ToM} region. Although DKT achieves higher scores on $\delta_{1.025}$ and $\delta_{1.05}$ in the \textbf{ToM} region owing to fine-tuning on the evaluation dataset, \modelname{} outperforms DKT on the remaining six metrics in a zero-shot setting. \modelname{} also outperforms DKT on the \textbf{All} and \textbf{Other} regions, where DKT's performance is limited by a tendency to overestimate depth in background areas. For a direct zero-shot comparison with DKT on an unseen dataset, we additionally evaluate on TDoF20~\cite{s26030898} (see Appendix and Tab.~\ref{tab:supp-tdof20}), where \modelname{} outperforms DKT on seven of the eight metrics, with iRMSE tied. For both Depth Anything V3 and MoGe-2, generative opacification consistently improves depth estimation performance. With Depth Anything V3, RMSE decreases from 24.03\,mm to 21.45\,mm (10.7\%). The improvement is more pronounced for MoGe-2, where RMSE is reduced from 56.83\,mm to 26.62\,mm and $\delta_{1.025}$ increases from 5.63\% to 42.75\%. These results suggest that replacing the transparent appearance with a generated opaque counterpart allows depth models to better estimate depth on transparent objects. 

Compared with specialized transparent-object depth methods not fine-tuned on the evaluation data, \modelname{} with Depth Anything V3 achieves lower RMSE than Depth4ToM (24.83\,mm), D4RD (39.95\,mm), and MODEST (39.31\,mm). It also outperforms diffusion-based depth models including GenPercept (24.09\,mm), Marigold (28.10\,mm), and GeoWizard (31.54\,mm), despite keeping the backbone frozen. Beyond RMSE, our method also improves other error metrics such as AbsRel and SiLog while maintaining competitive threshold accuracies. On the \textbf{All} and \textbf{Other} regions, \modelname{} with MoGe-2 achieves the best performance across all metrics, indicating that improvements on transparent regions do not degrade predictions elsewhere. Similarly, \modelname{} with Depth Anything V3 improves over the backbone baseline on most metrics, with the largest gain observed in RMSE (38.88\,mm to 34.85\,mm).

\textbf{Quantitative Results on TransPhy3D.} As shown in Tab.~\ref{tab:transphy3d}, \modelname{} achieves consistent improvements on the synthetic TransPhy3D benchmark. In particular, \modelname{} with MoGe-2 attains the best performance on six of the eight metrics and improves over the MoGe-2 baseline across all metrics. The largest gains compared to MoGe-2 are observed in AbsRel, which decreases from 0.016 to 0.012 (25\%), and in $\delta_{1.025}$, which increases from 89.58\% to 94.43\%, reflecting more pixels whose predicted depth closely match the ground truth. Similarly, \modelname{} with Depth Anything V3 consistently improves over the Depth Anything V3 across all eight metrics, reducing $\delta_{1.025}$ from 76.00\% to 88.01\% and RMSE from 53.25\,mm to 47.46\,mm. These results further demonstrate that appearance opacification enables general-purpose depth models to better handle transparent objects without requiring dataset-specific fine-tuning, offering stable and accurate depth estimation for complex transparent objects. Compared with specialized transparent-object depth methods, \modelname{} with MoGe-2 achieves comparable or better performance to DKT (fine-tuned on this dataset) across most metrics without any dataset-specific training—surpassing DKT on AbsRel (0.012 vs. 0.014) while remaining competitive on RMSE and threshold accuracies. It also outperforms D4RD (AbsRel 0.017), Depth4ToM (0.027), and MODEST (0.035), as well as general diffusion-based depth models including Marigold (0.027), GeoWizard (0.025), and GenPercept (0.037).

% ----------------------------------------------------------
\subsection{Qualitative Evaluation}
\label{sec:qual}
% ----------------------------------------------------------

\textbf{Qualitative Comparisons on Benchmark Datasets.}
Fig.~\ref{fig:comparison} presents qualitative comparisons with the baseline in the same test set in Sec. \ref{sec:quant}. General depth models such as Depth Anything V3 and MoGe-2 produce accurate depth in non-transparent regions but fail on transparent surfaces, which are often predicted as translucent or background-like due to the lack of reliable visual cues. Depth4ToM tends to produce solid but overly smooth depth predictions, resulting in vague object boundaries and reduced geometric detail. Our zero-shot method produces visually comparable transparent-object depth to DKT, which requires fine-tuning on this dataset. MODEST and D4RD often generate vague depth estimates, with interior regions appearing overly bright or translucent. GenPercept struggles with boundary accuracy and complex transparent structures, frequently producing translucent predictions and distorted depth near object edges. GeoWizard tends to generate piecewise or discretized depth with abrupt changes, leading to non-smooth geometry. Similarly, Marigold often produces discontinuous depth transitions and translucent artifacts. In contrast, our method produces stable and accurate depth predictions for complex transparent objects by generatively converting transparent appearances into geometry-consistent opaque representations. For instance, for an open jam bottle, our method accurately predicts the normalized depth map (circled in Fig.~\ref{fig:comparison}), where the opening appears lighter due to the hollow region, while the front bottle body appears darker because it is closer to the camera.

\textbf{Qualitative Comparisons on In-the-Wild Images.}
We compare \modelname{}+MoGe-2 against Depth Anything~V3, Depth4ToM, GenPercept, Marigold, and MoGe-2 on in-the-wild images across two figure sets (Fig.~\ref{fig:wild-1} and Fig.~\ref{fig:wild-2}). Scenes span a diverse range of challenges, including glasses, transparent objects with liquid and inter-object occlusion, out-of-distribution transparent containers enclosing opaque contents (e.g., a chocolate box with plastic film, a plastic drawer cabinet, a storage box), mixed transparent–opaque materials, and scenes with plants partially extending into and out of transparent vessels. Depth4ToM consistently produces blurrier boundaries than other methods; it also exhibits translucency on certain scenes and fails to reflect fine surface geometry, such as cup openings and the surface shape of bottles and containers. Depth Anything~V3 and MoGe-2 show pronounced translucency throughout both figure sets. Marigold and GenPercept degrade similarly: both exhibit translucency across a majority of scenes; In Fig.~\ref{fig:wild-2}, both Marigold and GenPercept additionally suffer from uneven depth transitions and produce depth maps that do not match object geometry (e.g., missing cup openings). In contrast, \modelname{} produces no translucency across all scenes, accurately recovers surface geometry including openings and contours, and maintains seamless depth transitions at material boundaries.

% ----------------------------------------------------------
\subsection{Ablation Study}
\label{sec:ablation}

% ----------------------------------------------------------

We conduct an ablation study on the key components of the proposed framework. 

\textbf{Transparent Object Localization.} To localize transparent objects, we use the mask $M^{seg}$ as spatial conditioning for generating opaque images. We compare two alternatives: bounding box, and point heatmap inputs. As shown in Tab.~\ref{tab:ablation}, the mask $M^{seg}$ consistently outperforms coarser cues. Using a bounding box increases RMSE from 21.45\,mm to 22.96\,mm, while a point heatmap yields 22.38\,mm, since imprecise spatial guidance leads to over- or under-opacification near object boundaries. As illustrated in Fig.~\ref{fig:ablation2}, bounding-box inputs provide correct localization but lack precise boundary delineation, resulting in inaccurate object contours; point heatmaps additionally suffer from ambiguous spatial extent, leading to unintended edits in surrounding regions.

\textbf{Image Encoder.}
We ablate three variants: CLIP CLS tokens, CLIP full tokens, DINOv2 CLS tokens, and DINOv2 full tokens. As shown in Tab.~\ref{tab:ablation}, the CLIP class token performs best (RMSE 21.45\,mm), while all alternatives degrade performance. As illustrated in Fig.~\ref{fig:ablation1}, weaker conditioning tokens lead to incomplete opacification, where regions near the object boundary or interior remain translucent, causing the depth model to produce translucent depth artifacts in those areas.

\textbf{Image Compositing.}
We explore a mask refinement module with alpha blending to composite the opacified region from the generated image $I_{pred}$ with the original background $I_{tr}$, preserving scene context. Without MRM, segmentation errors cause incomplete opacification near edges, increasing iRMSE from 0.007 to 0.008. Without blending, compositing is determined solely by patch boundaries; in multi-object scenes, a region shared by multiple patches may be overwritten by a later patch that treats it as background, restoring the original transparent appearance. As shown in Fig.~\ref{fig:ablation3}, this causes translucent depth artifacts in the overlapping region. This raises RMSE to 23.60\,mm and reduces $\delta{<}1.025$ to 45.45\%.

\textbf{Perceptual Loss.} We ablate the choice of auxiliary pixel-space loss added to the latent diffusion model objective $\mathcal{L}_{\text{LDM}}$; formal definitions of the alternative losses are provided in the Appendix. As shown in Tab.~\ref{tab:ablation} and Fig.~\ref{fig:ablation-loss}, training with $\mathcal{L}_{\text{LDM}}$ performs worst on dense geometric metrics, with RMSE 22.30\,mm. $\mathcal{L}_{\mathrm{grad}}$ yields only marginal improvement (RMSE 22.21\,mm), as transparent surfaces frequently exhibit optical effects such as caustics and specular highlights, which produce sharp luminance gradients unrelated to surface geometry. $\mathcal{L}_{\mathrm{L1}}$ reduces RMSE to 21.81\,mm but MAE remains higher than our method (18.19\,mm vs.\ 17.94\,mm), as uniform pixel-level supervision is sensitive to colour drift and does not distinguish perceptually important structures from flat background regions. Our $\mathcal{L}_{\mathrm{LPIPS}}$ variant achieves the best RMSE (21.45\,mm) and MAE (17.94\,mm), since feature-level perceptual supervision captures multi-scale structural cues that are robust to localised luminance artifacts.

\textbf{Opacified Image.} Our method employs a generative opacification module to convert transparent regions into shading- and geometry-consistent opaque representations. To evaluate its effect, we replace the generated opaque image with a flat-filled baseline and feed the original input directly to the depth model. Following the solid-color inpainting strategy of Depth4ToM~\cite{costanzino2023learning}, we fill the transparent region with red, green, and blue colors respectively and take the pixel-wise median of the three depth predictions, using the same $M^{seg}$. This baseline yields the largest degradation across all ablations (RMSE 33.09\,mm), falling below even the unmodified DA3 baseline (24.03\,mm): flat fills carry no structural cues, and mask inaccuracies produce incorrect object boundaries that mislead the depth backbone. While DA3 itself produces translucent depth artifacts on transparent regions, our diffusion-generated opaque images restore physically plausible surface appearance, enabling accurate depth estimation.
%%%%%%%%%%%%%%%%%%%%%%%%%%%%%%%%%%%%%%%%%%%%%%%%%%%%%%%%%%%%
% Section 5: Conclusion
%%%%%%%%%%%%%%%%%%%%%%%%%%%%%%%%%%%%%%%%%%%%%%%%%%%%%%%%%%%%
\section{Conclusion}
\label{sec:conclusion}
We presented \modelname{}, a novel two-stage framework for accurate depth estimation of transparent objects from a single image. By integrating a plug-and-play generative opacification module with mask-aware refinement and off-the-shelf monocular depth estimation, \modelname{} decouples appearance opacification and depth inference. By converting transparent appearances into geometry-consistent opaque appearance, the framework removes refractive and transmissive effects that lie outside the training distribution of foundation depth models, enabling foundation depth models to produce stable and accurate predictions without retraining. In future work, we plan to extend \modelname{} to video-based depth estimation and more complex non-Lambertian materials, including specular, translucent, and multi-layer transparent objects. We also aim to incorporate temporal consistency and spatial coherence constraints to ensure stable predictions across frames and improved robustness under occlusions.

{
    \small
    \FloatBarrier
    \bibliographystyle{ieeenat_fullname}
    \bibliography{main}

\begin{thebibliography}{68}
\providecommand{\natexlab}[1]{#1}
\providecommand{\url}[1]{\texttt{#1}}
\expandafter\ifx\csname urlstyle\endcsname\relax
  \providecommand{\doi}[1]{doi: #1}\else
  \providecommand{\doi}{doi: \begingroup \urlstyle{rm}\Url}\fi

\bibitem[Bhat et~al.(2023)Bhat, Birkl, Wofk, Wonka, and Müller]{bhat2023zoedepthzeroshottransfercombining}
Shariq~Farooq Bhat, Reiner Birkl, Diana Wofk, Peter Wonka, and Matthias Müller.
\newblock Zoedepth: Zero-shot transfer by combining relative and metric depth, 2023.

\bibitem[Birkl et~al.(2023)Birkl, Wofk, and M{\"u}ller]{birkl2023midas}
Reiner Birkl, Diana Wofk, and Matthias M{\"u}ller.
\newblock Midas v3. 1--a model zoo for robust monocular relative depth estimation.
\newblock \emph{arXiv preprint arXiv:2307.14460}, 2023.

\bibitem[Bradbury and Zhong(2025)]{bradbury2025latentmaskwrongpixelequivalent}
Rowan Bradbury and Dazhi Zhong.
\newblock Your latent mask is wrong: Pixel-equivalent latent compositing for diffusion models, 2025.

\bibitem[Brooks et~al.(2023)Brooks, Holynski, and Efros]{brooks2023instructpix2pix}
Tim Brooks, Aleksander Holynski, and Alexei~A Efros.
\newblock Instructpix2pix: Learning to follow image editing instructions.
\newblock In \emph{Proceedings of the IEEE/CVF conference on computer vision and pattern recognition}, pages 18392--18402, 2023.

\bibitem[Careaga and Aksoy(2024)]{Careaga_2024}
Chris Careaga and Yağız Aksoy.
\newblock Colorful diffuse intrinsic image decomposition in the wild.
\newblock \emph{ACM Transactions on Graphics}, 43\penalty0 (6):\penalty0 1–12, 2024.

\bibitem[Carion et~al.(2025)Carion, Gustafson, Hu, Debnath, Hu, Suris, Ryali, Alwala, Khedr, Huang, Lei, Ma, Guo, Kalla, Marks, Greer, Wang, Sun, Rädle, Afouras, Mavroudi, Xu, Wu, Zhou, Momeni, Hazra, Ding, Vaze, Porcher, Li, Li, Kamath, Cheng, Dollár, Ravi, Saenko, Zhang, and Feichtenhofer]{carion2025sam3segmentconcepts}
Nicolas Carion, Laura Gustafson, Yuan-Ting Hu, Shoubhik Debnath, Ronghang Hu, Didac Suris, Chaitanya Ryali, Kalyan~Vasudev Alwala, Haitham Khedr, Andrew Huang, Jie Lei, Tengyu Ma, Baishan Guo, Arpit Kalla, Markus Marks, Joseph Greer, Meng Wang, Peize Sun, Roman Rädle, Triantafyllos Afouras, Effrosyni Mavroudi, Katherine Xu, Tsung-Han Wu, Yu Zhou, Liliane Momeni, Rishi Hazra, Shuangrui Ding, Sagar Vaze, Francois Porcher, Feng Li, Siyuan Li, Aishwarya Kamath, Ho~Kei Cheng, Piotr Dollár, Nikhila Ravi, Kate Saenko, Pengchuan Zhang, and Christoph Feichtenhofer.
\newblock Sam 3: Segment anything with concepts, 2025.

\bibitem[Chen et~al.(2022)Chen, Zhang, Yu, Opipari, and Chadwicke~Jenkins]{chen2022clearpose}
Xiaotong Chen, Huijie Zhang, Zeren Yu, Anthony Opipari, and Odest Chadwicke~Jenkins.
\newblock Clearpose: Large-scale transparent object dataset and benchmark.
\newblock In \emph{European conference on computer vision}, pages 381--396. Springer, 2022.

\bibitem[Chen et~al.(2024)Chen, Huang, Liu, Shen, Zhao, and Zhao]{chen2024anydoor}
Xi Chen, Lianghua Huang, Yu Liu, Yujun Shen, Deli Zhao, and Hengshuang Zhao.
\newblock Anydoor: Zero-shot object-level image customization.
\newblock In \emph{Proceedings of the IEEE/CVF conference on computer vision and pattern recognition}, pages 6593--6602, 2024.

\bibitem[Costanzino et~al.(2023)Costanzino, Ramirez, Poggi, Tosi, Mattoccia, and Di~Stefano]{costanzino2023learning}
Alex Costanzino, Pierluigi~Zama Ramirez, Matteo Poggi, Fabio Tosi, Stefano Mattoccia, and Luigi Di~Stefano.
\newblock Learning depth estimation for transparent and mirror surfaces.
\newblock In \emph{Proceedings of the IEEE/CVF International Conference on Computer Vision}, pages 9244--9255, 2023.

\bibitem[Eigen et~al.(2014)Eigen, Puhrsch, and Fergus]{eigen2014depthmappredictionsingle}
David Eigen, Christian Puhrsch, and Rob Fergus.
\newblock Depth map prediction from a single image using a multi-scale deep network, 2014.

\bibitem[Fang et~al.(2022)Fang, Fang, Xu, and Lu]{fang2022transcg}
Hongjie Fang, Hao-Shu Fang, Sheng Xu, and Cewu Lu.
\newblock Transcg: A large-scale real-world dataset for transparent object depth completion and a grasping baseline.
\newblock \emph{IEEE Robotics and Automation Letters}, pages 1--8, 2022.

\bibitem[Flacco et~al.(2012)Flacco, Kr{\"o}ger, De~Luca, and Khatib]{flacco2012depth}
Fabrizio Flacco, Torsten Kr{\"o}ger, Alessandro De~Luca, and Oussama Khatib.
\newblock A depth space approach to human-robot collision avoidance.
\newblock In \emph{2012 IEEE international conference on robotics and automation}, pages 338--345. IEEE, 2012.

\bibitem[Flacco et~al.(2015)Flacco, Kroeger, De~Luca, and Khatib]{flacco2015depth}
Fabrizio Flacco, Torsten Kroeger, Alessandro De~Luca, and Oussama Khatib.
\newblock A depth space approach for evaluating distance to objects: with application to human-robot collision avoidance.
\newblock \emph{Journal of Intelligent \& Robotic Systems}, 80\penalty0 (Suppl 1):\penalty0 7--22, 2015.

\bibitem[Fu et~al.(2024)Fu, Yin, Hu, Wang, Ma, Tan, Shen, Lin, and Long]{fu2024geowizard}
Xiao Fu, Wei Yin, Mu Hu, Kaixuan Wang, Yuexin Ma, Ping Tan, Shaojie Shen, Dahua Lin, and Xiaoxiao Long.
\newblock Geowizard: Unleashing the diffusion priors for 3d geometry estimation from a single image.
\newblock In \emph{European Conference on Computer Vision}, pages 241--258. Springer, 2024.

\bibitem[Guo et~al.(2025)Guo, Zhu, Peng, Lin, Yan, Xie, Wang, Zhou, and Bao]{guo2025murre}
Haoyu Guo, He Zhu, Sida Peng, Haotong Lin, Yunzhi Yan, Tao Xie, Wenguan Wang, Xiaowei Zhou, and Hujun Bao.
\newblock Multi-view reconstruction via sfm-guided monocular depth estimation.
\newblock In \emph{CVPR}, 2025.

\bibitem[Ho et~al.(2020)Ho, Jain, and Abbeel]{ho2020denoising}
Jonathan Ho, Ajay Jain, and Pieter Abbeel.
\newblock Denoising diffusion probabilistic models.
\newblock \emph{Advances in neural information processing systems}, 33:\penalty0 6840--6851, 2020.

\bibitem[Hodan et~al.()]{hodan2020bop}
Tomas Hodan et~al.
\newblock Bop challenge 2020 on 6d object localization.
\newblock In \emph{ECCV Workshops}.

\bibitem[Kalra et~al.(2020)Kalra, Taamazyan, Rao, Venkataraman, Raskar, and Kadambi]{Kalra_2020_CVPR}
Agastya Kalra, Vage Taamazyan, Supreeth~Krishna Rao, Kartik Venkataraman, Ramesh Raskar, and Achuta Kadambi.
\newblock Deep polarization cues for transparent object segmentation.
\newblock In \emph{Proceedings of the IEEE/CVF Conference on Computer Vision and Pattern Recognition (CVPR)}, 2020.

\bibitem[Ke et~al.(2024)Ke, Obukhov, Huang, Metzger, Daudt, and Schindler]{ke2024repurposingdiffusionbasedimagegenerators}
Bingxin Ke, Anton Obukhov, Shengyu Huang, Nando Metzger, Rodrigo~Caye Daudt, and Konrad Schindler.
\newblock Repurposing diffusion-based image generators for monocular depth estimation, 2024.

\bibitem[Ke et~al.(2025)Ke, Qu, Wang, Metzger, Huang, Li, Obukhov, and Schindler]{ke2025marigold}
Bingxin Ke, Kevin Qu, Tianfu Wang, Nando Metzger, Shengyu Huang, Bo Li, Anton Obukhov, and Konrad Schindler.
\newblock Marigold: Affordable adaptation of diffusion-based image generators for image analysis.
\newblock \emph{IEEE Transactions on Pattern Analysis and Machine Intelligence}, 2025.

\bibitem[Keetha et~al.(2025)Keetha, M{\"u}ller, Sch{\"o}nberger, Porzi, Zhang, Fischer, Knapitsch, Zauss, Weber, Antunes, et~al.]{keetha2025mapanything}
Nikhil Keetha, Norman M{\"u}ller, Johannes Sch{\"o}nberger, Lorenzo Porzi, Yuchen Zhang, Tobias Fischer, Arno Knapitsch, Duncan Zauss, Ethan Weber, Nelson Antunes, et~al.
\newblock Mapanything: Universal feed-forward metric 3d reconstruction.
\newblock \emph{arXiv preprint arXiv:2509.13414}, 2025.

\bibitem[Kirillov et~al.(2023)Kirillov, Mintun, Ravi, Mao, Rolland, Gustafson, Xiao, Whitehead, Berg, Lo, Dollár, and Girshick]{kirillov2023segment}
Alexander Kirillov, Eric Mintun, Nikhila Ravi, Hanzi Mao, Chloe Rolland, Laura Gustafson, Tete Xiao, Spencer Whitehead, Alexander~C. Berg, Wan-Yen Lo, Piotr Dollár, and Ross Girshick.
\newblock Segment anything, 2023.

\bibitem[Lee et~al.(2020)Lee, Lee, Oh, and Kweon]{lee2020cnn}
Byeong-Uk Lee, Kyunghyun Lee, Jean Oh, and In~So Kweon.
\newblock Cnn-based simultaneous dehazing and depth estimation.
\newblock In \emph{2020 IEEE international conference on robotics and automation (ICRA)}, pages 9722--9728. IEEE, 2020.

\bibitem[Li et~al.(2024)Li, Wang, Peng, Shen, Zhu, and Xu]{li2024visual}
Jinming Li, Wanying Wang, Yaxin Peng, Chaomin Shen, Yichen Zhu, and Zhiyuan Xu.
\newblock Visual robotic manipulation with depth-aware pretraining.
\newblock In \emph{2024 IEEE International Conference on Robotics and Biomimetics (ROBIO)}, pages 843--850. IEEE, 2024.

\bibitem[Lin et~al.(2025)Lin, Chen, Liew, Chen, Li, Shi, Feng, and Kang]{lin2025depth}
Haotong Lin, Sili Chen, Junhao Liew, Donny~Y Chen, Zhenyu Li, Guang Shi, Jiashi Feng, and Bingyi Kang.
\newblock Depth anything 3: Recovering the visual space from any views.
\newblock \emph{arXiv preprint arXiv:2511.10647}, 2025.

\bibitem[Liu et~al.(2017)Liu, Ceylan, Yumer, Yang, and Lien]{liu2017material}
Guilin Liu, Duygu Ceylan, Ersin Yumer, Jimei Yang, and Jyh-Ming Lien.
\newblock Material editing using a physically based rendering network.
\newblock In \emph{Proceedings of the IEEE International Conference on Computer Vision}, pages 2261--2269, 2017.

\bibitem[Liu et~al.(2025)Liu, Ma, Guo, Zhao, Zhang, Sui, and Zou]{liu2025monocular}
Jiangyuan Liu, Hongxuan Ma, Yuxin Guo, Yuhao Zhao, Chi Zhang, Wei Sui, and Wei Zou.
\newblock Monocular depth estimation and segmentation for transparent object with iterative semantic and geometric fusion.
\newblock \emph{arXiv preprint arXiv:2502.14616}, 2025.

\bibitem[Ma et~al.(2025{\natexlab{a}})Ma, Ma, Li, Zheng, and Li]{ma2025tosq}
Bin Ma, Ming Ma, Ruiguang Li, Jiawei Zheng, and Deping Li.
\newblock Tosq: Transparent object segmentation via query-based dictionary lookup with transformers.
\newblock \emph{Sensors}, 25\penalty0 (15):\penalty0 4700, 2025{\natexlab{a}}.

\bibitem[Ma et~al.(2025{\natexlab{b}})Ma, Fang, Weibel, Tan, Wang, Xiao, Fang, and Xia]{ma2025physliquidphysicsinformeddatasetestimating}
Ke Ma, Yizhou Fang, Jean-Baptiste Weibel, Shuai Tan, Xinggang Wang, Yang Xiao, Yi Fang, and Tian Xia.
\newblock Phys-liquid: A physics-informed dataset for estimating 3d geometry and volume of transparent deformable liquids, 2025{\natexlab{b}}.

\bibitem[Ma et~al.(2026)Ma, Xu, and Zhang]{ma2026pixelgen}
Zehong Ma, Ruihan Xu, and Shiliang Zhang.
\newblock Pixelgen: Pixel diffusion beats latent diffusion with perceptual loss.
\newblock \emph{arXiv preprint arXiv:2602.02493}, 2026.

\bibitem[Mahler et~al.(2017)Mahler, Liang, Niyaz, Laskey, Doan, Liu, Ojea, and Goldberg]{mahler2017dexnet20deeplearning}
Jeffrey Mahler, Jacky Liang, Sherdil Niyaz, Michael Laskey, Richard Doan, Xinyu Liu, Juan~Aparicio Ojea, and Ken Goldberg.
\newblock Dex-net 2.0: Deep learning to plan robust grasps with synthetic point clouds and analytic grasp metrics, 2017.

\bibitem[Nathan~Silberman and Fergus(2012)]{Silberman:ECCV12}
Pushmeet~Kohli Nathan~Silberman, Derek~Hoiem and Rob Fergus.
\newblock Indoor segmentation and support inference from rgbd images.
\newblock In \emph{ECCV}, 2012.

\bibitem[Newcombe et~al.()Newcombe, Izadi, et~al.]{newcombe2011kinectfusion}
Richard~A. Newcombe, Shahram Izadi, et~al.
\newblock Kinectfusion: Real-time dense surface mapping and tracking.
\newblock In \emph{IEEE Int. Symp. on Mixed and Augmented Reality (ISMAR)}.

\bibitem[Oquab et~al.(2024)Oquab, Darcet, Moutakanni, Vo, Szafraniec, Khalidov, Fernandez, Haziza, Massa, El-Nouby, Assran, Ballas, Galuba, Howes, Huang, Li, Misra, Rabbat, Sharma, Synnaeve, Xu, Jegou, Mairal, Labatut, Joulin, and Bojanowski]{oquab2024dinov2learningrobustvisual}
Maxime Oquab, Timothée Darcet, Théo Moutakanni, Huy Vo, Marc Szafraniec, Vasil Khalidov, Pierre Fernandez, Daniel Haziza, Francisco Massa, Alaaeldin El-Nouby, Mahmoud Assran, Nicolas Ballas, Wojciech Galuba, Russell Howes, Po-Yao Huang, Shang-Wen Li, Ishan Misra, Michael Rabbat, Vasu Sharma, Gabriel Synnaeve, Hu Xu, Hervé Jegou, Julien Mairal, Patrick Labatut, Armand Joulin, and Piotr Bojanowski.
\newblock Dinov2: Learning robust visual features without supervision, 2024.

\bibitem[Qiu et~al.(2025)Qiu, Du, Spine, Cheng, and Jiang]{qiu2025joint3dpointcloud}
Tian Qiu, Ruiming Du, Nikolai Spine, Lailiang Cheng, and Yu Jiang.
\newblock Joint 3d point cloud segmentation using real-sim loop: From panels to trees and branches, 2025.

\bibitem[Radford et~al.(2021)Radford, Kim, Hallacy, Ramesh, Goh, Agarwal, Sastry, Askell, Mishkin, Clark, et~al.]{radford2021learning}
Alec Radford, Jong~Wook Kim, Chris Hallacy, Aditya Ramesh, Gabriel Goh, Sandhini Agarwal, Girish Sastry, Amanda Askell, Pamela Mishkin, Jack Clark, et~al.
\newblock Learning transferable visual models from natural language supervision.
\newblock In \emph{International conference on machine learning}, pages 8748--8763. PmLR, 2021.

\bibitem[Ramirez et~al.(2024)Ramirez, Costanzino, Tosi, Poggi, Salti, Mattoccia, and Stefano]{ramirez2024boosterbenchmarkdepthimages}
Pierluigi~Zama Ramirez, Alex Costanzino, Fabio Tosi, Matteo Poggi, Samuele Salti, Stefano Mattoccia, and Luigi~Di Stefano.
\newblock Booster: a benchmark for depth from images of specular and transparent surfaces, 2024.

\bibitem[Ranftl et~al.(2020)Ranftl, Lasinger, Hafner, Schindler, and Koltun]{ranftl2020towards}
Ren{\'e} Ranftl, Katrin Lasinger, David Hafner, Konrad Schindler, and Vladlen Koltun.
\newblock Towards robust monocular depth estimation: Mixing datasets for zero-shot cross-dataset transfer.
\newblock \emph{IEEE transactions on pattern analysis and machine intelligence}, 44\penalty0 (3):\penalty0 1623--1637, 2020.

\bibitem[Ranftl et~al.(2021)Ranftl, Bochkovskiy, and Koltun]{ranftl2021visiontransformersdenseprediction}
René Ranftl, Alexey Bochkovskiy, and Vladlen Koltun.
\newblock Vision transformers for dense prediction, 2021.

\bibitem[Ren et~al.(2024)Ren, Liu, Zeng, Lin, Li, Cao, Chen, Huang, Chen, Yan, Zeng, Zhang, Li, Yang, Li, Jiang, and Zhang]{ren2024groundedsamassemblingopenworld}
Tianhe Ren, Shilong Liu, Ailing Zeng, Jing Lin, Kunchang Li, He Cao, Jiayu Chen, Xinyu Huang, Yukang Chen, Feng Yan, Zhaoyang Zeng, Hao Zhang, Feng Li, Jie Yang, Hongyang Li, Qing Jiang, and Lei Zhang.
\newblock Grounded sam: Assembling open-world models for diverse visual tasks, 2024.

\bibitem[Rombach et~al.(2022)Rombach, Blattmann, Lorenz, Esser, and Ommer]{rombach2022high}
Robin Rombach, Andreas Blattmann, Dominik Lorenz, Patrick Esser, and Bj{\"o}rn Ommer.
\newblock High-resolution image synthesis with latent diffusion models.
\newblock In \emph{Proceedings of the IEEE/CVF conference on computer vision and pattern recognition}, pages 10684--10695, 2022.

\bibitem[Sajjan et~al.(2020)Sajjan, Moore, Pan, Nagaraja, Lee, Zeng, and Song]{sajjan2020clear}
Shreeyak Sajjan, Matthew Moore, Mike Pan, Ganesh Nagaraja, Johnny Lee, Andy Zeng, and Shuran Song.
\newblock Clear grasp: 3d shape estimation of transparent objects for manipulation.
\newblock In \emph{2020 IEEE international conference on robotics and automation (ICRA)}, pages 3634--3642. IEEE, 2020.

\bibitem[Sajjan et~al.()]{sajjan2020cleargrasp}
Shreeyak~S. Sajjan et~al.
\newblock Cleargrasp: 3d shape estimation of transparent objects for manipulation.
\newblock In \emph{ICRA}.

\bibitem[Schmidt et~al.(2015)Schmidt, Hertkorn, Newcombe, Marton, Suppa, and Fox]{schmidt2015depth}
Tanner Schmidt, Katharina Hertkorn, Richard Newcombe, Zoltan Marton, Michael Suppa, and Dieter Fox.
\newblock Depth-based tracking with physical constraints for robot manipulation.
\newblock In \emph{2015 IEEE International Conference on Robotics and Automation (ICRA)}, pages 119--126. IEEE, 2015.

\bibitem[Sch{\"o}nberger and Frahm()]{schonberger2016sfm}
Johannes~L. Sch{\"o}nberger and Jan-Michael Frahm.
\newblock Structure-from-motion revisited.
\newblock In \emph{CVPR}.

\bibitem[Shankar et~al.(2022)Shankar, Tjersland, Ma, Stone, and Bajracharya]{shankar2022learned}
Krishna Shankar, Mark Tjersland, Jeremy Ma, Kevin Stone, and Max Bajracharya.
\newblock A learned stereo depth system for robotic manipulation in homes.
\newblock \emph{IEEE Robotics and Automation Letters}, 7\penalty0 (2):\penalty0 2305--2312, 2022.

\bibitem[Sharma et~al.(2023)Sharma, Jampani, Li, Jia, Lagun, Durand, Freeman, and Matthews]{sharma2023alchemistparametriccontrolmaterial}
Prafull Sharma, Varun Jampani, Yuanzhen Li, Xuhui Jia, Dmitry Lagun, Fredo Durand, William~T. Freeman, and Mark Matthews.
\newblock Alchemist: Parametric control of material properties with diffusion models, 2023.

\bibitem[Silberman and Fergus(2011)]{silberman11indoor}
N. Silberman and R. Fergus.
\newblock Indoor scene segmentation using a structured light sensor.
\newblock In \emph{Proceedings of the International Conference on Computer Vision - Workshop on 3D Representation and Recognition}, 2011.

\bibitem[Sim{\'e}oni et~al.(2025)Sim{\'e}oni, Vo, Seitzer, Baldassarre, Oquab, Jose, Khalidov, Szafraniec, Yi, Ramamonjisoa, et~al.]{simeoni2025dinov3}
Oriane Sim{\'e}oni, Huy~V Vo, Maximilian Seitzer, Federico Baldassarre, Maxime Oquab, Cijo Jose, Vasil Khalidov, Marc Szafraniec, Seungeun Yi, Micha{\"e}l Ramamonjisoa, et~al.
\newblock Dinov3.
\newblock \emph{arXiv preprint arXiv:2508.10104}, 2025.

\bibitem[ten Pas et~al.()ten Pas, Gualtieri, Saenko, and Platt]{tenpas2017gpd}
Andreas ten Pas, Marcus Gualtieri, Kate Saenko, and Robert Platt.
\newblock Grasp pose detection in point clouds.

\bibitem[Tosi et~al.(2024)Tosi, {Zama Ramirez}, and Poggi]{tosi2024diffusion}
Fabio Tosi, Pierluigi {Zama Ramirez}, and Matteo Poggi.
\newblock Diffusion models for monocular depth estimation: {Overcoming} challenging conditions.
\newblock In \emph{European Conference on Computer Vision ({ECCV})}, 2024.

\bibitem[Wang et~al.(2025{\natexlab{a}})Wang, Tran, Cui, TG, Dahl, Bigdeli, Frisvad, and Chandraker]{wang2025materialistphysicallybasedediting}
Lezhong Wang, Duc~Minh Tran, Ruiqi Cui, Thomson TG, Anders~Bjorholm Dahl, Siavash~Arjomand Bigdeli, Jeppe~Revall Frisvad, and Manmohan Chandraker.
\newblock Materialist: Physically based editing using single-image inverse rendering, 2025{\natexlab{a}}.

\bibitem[Wang et~al.(2025{\natexlab{b}})Wang, Xu, Dong, Deng, Xiang, Lv, Sun, Tong, and Yang]{wang2025moge}
Ruicheng Wang, Sicheng Xu, Yue Dong, Yu Deng, Jianfeng Xiang, Zelong Lv, Guangzhong Sun, Xin Tong, and Jiaolong Yang.
\newblock Moge-2: Accurate monocular geometry with metric scale and sharp details.
\newblock \emph{arXiv preprint arXiv:2507.02546}, 2025{\natexlab{b}}.

\bibitem[Wang et~al.(2026)Wang, Wu, and Zou]{s26030898}
Yunhe Wang, Ting Wu, and Qin Zou.
\newblock 6dof pose estimation of transparent objects: Dataset and method.
\newblock \emph{Sensors}, 26\penalty0 (3), 2026.

\bibitem[Xie et~al.(2021{\natexlab{a}})Xie, Wang, Wang, Sun, Xu, Liang, and Luo]{xie2021segmenting}
Enze Xie, Wenjia Wang, Wenhai Wang, Peize Sun, Hang Xu, Ding Liang, and Ping Luo.
\newblock Segmenting transparent object in the wild with transformer.
\newblock \emph{arXiv preprint arXiv:2101.08461}, 2021{\natexlab{a}}.

\bibitem[Xie et~al.(2021{\natexlab{b}})Xie, Wang, Wang, Sun, Xu, Liang, and Luo]{xie2021segmentingtransparentobjectwild}
Enze Xie, Wenjia Wang, Wenhai Wang, Peize Sun, Hang Xu, Ding Liang, and Ping Luo.
\newblock Segmenting transparent object in the wild with transformer, 2021{\natexlab{b}}.

\bibitem[Xu et~al.(2024)Xu, Ge, Liu, Fan, Xie, Zhao, Chen, and Shen]{xu2024matters}
Guangkai Xu, Yongtao Ge, Mingyu Liu, Chengxiang Fan, Kangyang Xie, Zhiyue Zhao, Hao Chen, and Chunhua Shen.
\newblock What matters when repurposing diffusion models for general dense perception tasks?
\newblock \emph{arXiv preprint arXiv:2403.06090}, 2024.

\bibitem[Xu et~al.(2025{\natexlab{a}})Xu, Wei, Wei, Geng, Li, Shen, Sun, Han, Ma, Li, Ye, Zheng, Wang, Zhang, and Zhao]{dkt2025}
Shaocong Xu, Songlin Wei, Qizhe Wei, Zheng Geng, Hong Li, Licheng Shen, Qianpu Sun, Shu Han, Bin Ma, Bohan Li, Chongjie Ye, Yuhang Zheng, Nan Wang, Saining Zhang, and Hao Zhao.
\newblock Diffusion knows transparency: Repurposing video diffusion for transparent object depth and normal estimation.
\newblock \emph{https://arxiv.org/abs/2512.23705}, 2025{\natexlab{a}}.

\bibitem[Xu et~al.(2025{\natexlab{b}})Xu, Wei, Wei, Geng, Li, Shen, Sun, Han, Ma, Li, et~al.]{xu2025diffusion}
Shaocong Xu, Songlin Wei, Qizhe Wei, Zheng Geng, Hong Li, Licheng Shen, Qianpu Sun, Shu Han, Bin Ma, Bohan Li, et~al.
\newblock Diffusion knows transparency: Repurposing video diffusion for transparent object depth and normal estimation.
\newblock \emph{arXiv preprint arXiv:2512.23705}, 2025{\natexlab{b}}.

\bibitem[Yang et~al.(2023)Yang, Gu, Zhang, Zhang, Chen, Sun, Chen, and Wen]{yang2023paint}
Binxin Yang, Shuyang Gu, Bo Zhang, Ting Zhang, Xuejin Chen, Xiaoyan Sun, Dong Chen, and Fang Wen.
\newblock Paint by example: Exemplar-based image editing with diffusion models.
\newblock In \emph{Proceedings of the IEEE/CVF conference on computer vision and pattern recognition}, pages 18381--18391, 2023.

\bibitem[Yang et~al.(2024{\natexlab{a}})Yang, Kang, Huang, Xu, Feng, and Zhao]{yang2024depth}
Lihe Yang, Bingyi Kang, Zilong Huang, Xiaogang Xu, Jiashi Feng, and Hengshuang Zhao.
\newblock Depth anything: Unleashing the power of large-scale unlabeled data.
\newblock In \emph{Proceedings of the IEEE/CVF conference on computer vision and pattern recognition}, pages 10371--10381, 2024{\natexlab{a}}.

\bibitem[Yang et~al.(2024{\natexlab{b}})Yang, Kang, Huang, Zhao, Xu, Feng, and Zhao]{yang2024depthv2}
Lihe Yang, Bingyi Kang, Zilong Huang, Zhen Zhao, Xiaogang Xu, Jiashi Feng, and Hengshuang Zhao.
\newblock Depth anything v2, 2024{\natexlab{b}}.

\bibitem[Yang et~al.()]{yang2024depthanythingv2}
Lintu Yang et~al.
\newblock Depth anything v2.

\bibitem[Zhang et~al.(2023)Zhang, Wang, Ling, Guan, Zhang, Li, Wei, and Zhang]{zhang2023shuffletrans}
Boxiang Zhang, Zunran Wang, Yonggen Ling, Yuanyuan Guan, Shenghao Zhang, Wenhui Li, Lei Wei, and Chunxu Zhang.
\newblock Shuffletrans: Patch-wise weight shuffle for transparent object segmentation.
\newblock \emph{Neural Networks}, 167:\penalty0 199--212, 2023.

\bibitem[Zhang et~al.(2021)Zhang, Yang, Constantinescu, Peng, Müller, and Stiefelhagen]{zhang2021trans4transefficienttransformertransparent}
Jiaming Zhang, Kailun Yang, Angela Constantinescu, Kunyu Peng, Karin Müller, and Rainer Stiefelhagen.
\newblock Trans4trans: Efficient transformer for transparent object segmentation to help visually impaired people navigate in the real world, 2021.

\bibitem[Zhang et~al.(2018)Zhang, Isola, Efros, Shechtman, and Wang]{zhang2018unreasonable}
Richard Zhang, Phillip Isola, Alexei~A Efros, Eli Shechtman, and Oliver Wang.
\newblock The unreasonable effectiveness of deep features as a perceptual metric.
\newblock In \emph{Proceedings of the IEEE conference on computer vision and pattern recognition}, pages 586--595, 2018.

\bibitem[Zhao et~al.(2023)Zhao, Bai, Rao, Zhou, and Lu]{zhao2023unipc}
Wenliang Zhao, Lujia Bai, Yongming Rao, Jie Zhou, and Jiwen Lu.
\newblock Unipc: A unified predictor-corrector framework for fast sampling of diffusion models.
\newblock \emph{Advances in Neural Information Processing Systems}, 36:\penalty0 49842--49869, 2023.

\bibitem[Zhou et~al.(2017)Zhou, Zhao, Puig, Fidler, Barriuso, and Torralba]{zhou2017scene}
Bolei Zhou, Hang Zhao, Xavier Puig, Sanja Fidler, Adela Barriuso, and Antonio Torralba.
\newblock Scene parsing through ade20k dataset.
\newblock In \emph{Proceedings of the IEEE conference on computer vision and pattern recognition}, pages 633--641, 2017.

\end{thebibliography}
}

\clearpage
\appendix
\section*{\Large Appendix}
\section{Additional Evaluation on TDoF20}

To evaluate DKT~\cite{xu2025diffusion} and our method in a zero-shot setting, we additionally evaluate on TDoF20~\cite{s26030898}, a real-world RGB-D dataset covering 20 transparent objects in single- and multi-object scenes with heavy inter-object occlusion. We sample the test set at a stride of 7 to avoid redundant consecutive frames and report all metrics within the transparent-object mask region. We report AbsRel, SiLog, RMSE (mm), iRMSE ($\text{mm}^{-1}$), MAE (mm), and threshold accuracies $\delta_{1.025}$, $\delta_{1.05}$, $\delta_{1.10}$.

\begin{table}[!h]
\centering
\caption{\textbf{TDoF20.} Zero-shot comparison on the real-world TDoF20 test set (stride-7 sampling), evaluated within the transparent-object mask region. \colorbox{Palette5}{\phantom{x}}: best result.}
\label{tab:supp-tdof20}
\makebox[\linewidth][c]{\resizebox{1.15\linewidth}{!}{%
\renewcommand{\arraystretch}{1.2}
\setlength{\tabcolsep}{6pt}
\setlength{\aboverulesep}{0pt}
\setlength{\belowrulesep}{0pt}
\setlength{\extrarowheight}{0.4pt}
\begin{tabular}{>{\raggedright\arraybackslash}p{1.8cm} *{8}{>{\centering\arraybackslash}p{1.1cm}}}
\toprule
\textbf{Method} & AbsRel$\downarrow$ & SiLog$\downarrow$ & RMSE$\downarrow$ & iRMSE$\downarrow$ & MAE$\downarrow$ & $\delta_{1.025}$$\uparrow$ & $\delta_{1.05}$$\uparrow$ & $\delta_{1.10}$$\uparrow$ \\
\midrule
DKT~\cite{xu2025diffusion} & 0.069 & 0.038 & 51.39 & \cellcolor{Palette5}\textbf{0.0001} & 45.59 & 24.84 & 46.28 & 75.12 \\
\midrule
\rowcolor{OursColor}\textbf{Ours (DA3)} & \cellcolor{Palette5}\textbf{0.053} & \cellcolor{Palette5}\textbf{0.037} & \cellcolor{Palette5}\textbf{42.00} & \cellcolor{Palette5}\textbf{0.0001} & \cellcolor{Palette5}\textbf{36.03} & \cellcolor{Palette5}\textbf{29.66} & \cellcolor{Palette5}\textbf{53.92} & \cellcolor{Palette5}\textbf{83.87} \\
\bottomrule
\end{tabular}
}}
\end{table}

\textbf{Quantitative Results on TDoF20.} As shown in \cref{tab:supp-tdof20}, \modelname{} with Depth Anything V3 outperforms DKT on seven of the eight metrics in this zero-shot setting, with iRMSE tied. AbsRel decreases from 0.069 to 0.053, RMSE from 51.39\,mm to 42.00\,mm, and MAE from 45.59\,mm to 36.03\,mm. Threshold accuracies also improve consistently, with $\delta_{1.025}$ increasing from 24.84\% to 29.66\% and $\delta_{1.10}$ from 75.12\% to 83.87\%, demonstrating that \modelname{} generalizes to unseen real-world datasets without dataset-specific fine-tuning.

\label{sec:supp-tdof20}
\begin{figure}[!h]
    \centering
    \includegraphics[width=\linewidth]{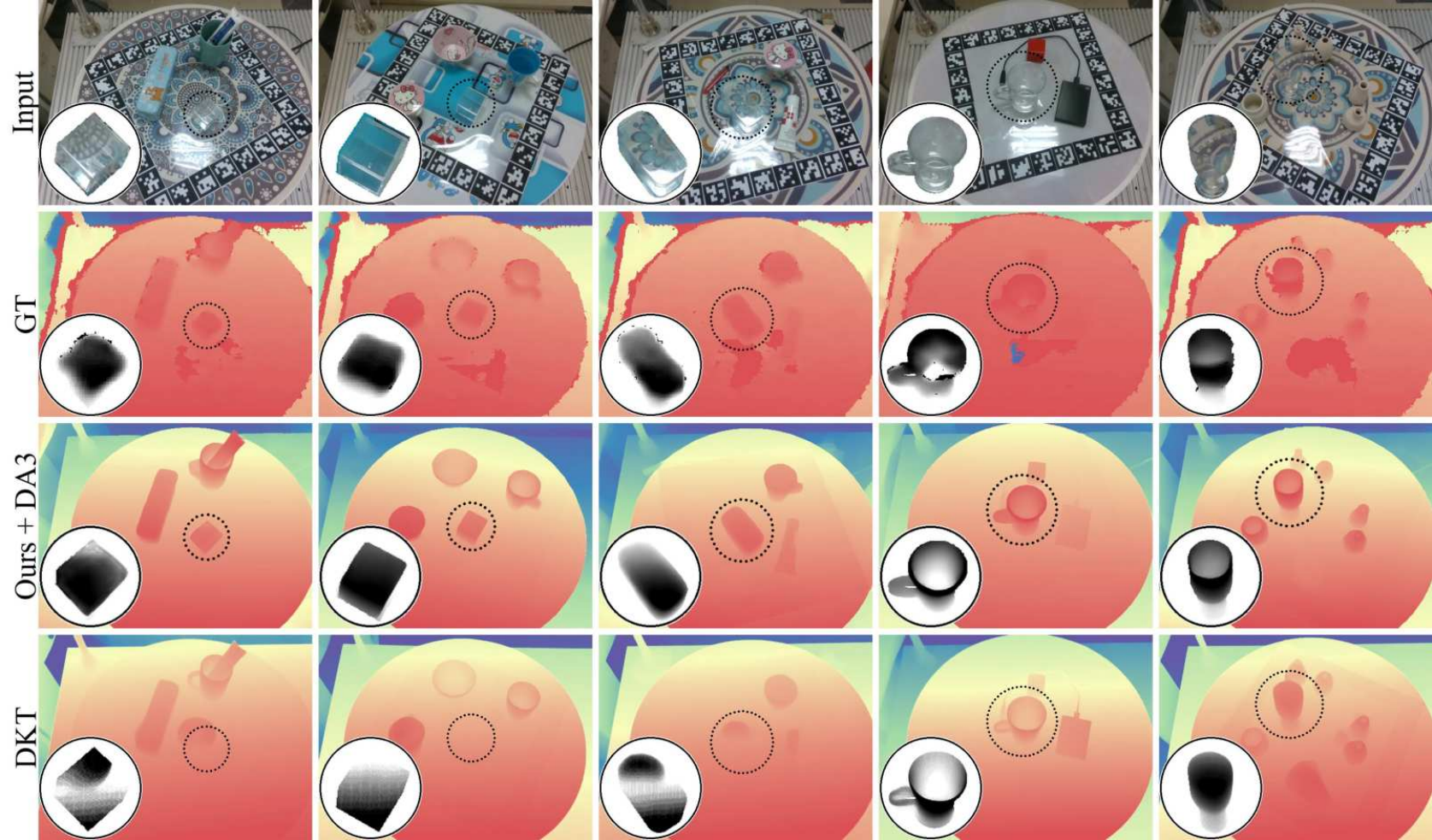}
    \caption{\textbf{Qualitative Comparison on TDoF20.} DKT produces inaccurate object boundaries within the transparent-object region and occasionally fails to recover the object structure entirely (dashed circle). Our method produces depth predictions consistent with the ground truth across all five scenes. The inset grayscale images show depth normalized within the transparent-object mask region; the intensity reflects the depth values within the mask and does not imply the presence of a solid object at that location. Depth maps in the circle are normalized in grayscale.}
    \label{fig:supp-tdof20-qual}
\end{figure}

\textbf{Qualitative Results on TDoF20.} As shown in \cref{fig:supp-tdof20-qual}, DKT produces inconsistent depth predictions across all five scenes. In column 1, the predicted object region is shifted toward the upper left relative to the actual object position. In column 2, the transparent object is entirely absent from the depth map. In column 3, only a small portion of the object is recovered, and the adjacent opaque object is also partially missing. In column 4, DKT overestimates the cup opening, produces a blurred handle, and fails to recover the thickness of the cup walls. In column 5, DKT predicts a solid depth surface for an open-top cup and introduces spurious depth variation on a specular highlight on the table surface below the object. In contrast, our method produces sharp, accurate boundaries and correctly recovers fine geometric details across all scenes, including cup wall thickness, opening size, and handle shape, consistent with both the ground truth and the input image.

%%%%%%%%%%%%%%%%%%%%%%%%%%%%%%%%%%%%%%%%%%%%%%%%%%%%%%%%%%%%
% Section D: Zero-Shot Comparison on TDoF20
%%%%%%%%%%%%%%%%%%%%%%%%%%%%%%%%%%%%%%%%%%%%%%%%%%%%%%%%%%%%

\section{Ablation Study: Perceptual Loss Variants}
\label{sec:supp-loss-ablation}

We define the two alternative auxiliary losses compared against $\mathcal{L}_{\mathrm{LPIPS}}$ in the main paper.

\textbf{Shading Gradient Loss.} A multi-scale Sobel gradient loss in log-luminance space over scales $\mathcal{S}=\{2,4,8\}$, masked to the morphologically eroded object interior:
\begin{equation}
\resizebox{0.88\linewidth}{!}{$\displaystyle
\mathcal{L}_{\mathrm{grad}} = \frac{1}{|\mathcal{S}|}\sum_{s\in\mathcal{S}} \frac{1}{\#M^{in}} \Bigl(
    \left\|\left(\nabla_x \log I^{pred}_s - \nabla_x \log I^{op}_s\right) \odot M^{in}\right\|_1
    + \left\|\left(\nabla_y \log I^{pred}_s - \nabla_y \log I^{op}_s\right) \odot M^{in}\right\|_1 \Bigr)
$}
\end{equation}
where $I^{pred}_s$ and $I^{op}_s$ are the predicted and ground-truth opaque images Gaussian-smoothed at scale $s$, $\nabla_x$ and $\nabla_y$ denote the Sobel gradient operators in the horizontal and vertical directions respectively, $M^{in} \in \{0,1\}^{H \times W}$ is the morphologically eroded interior mask of $M^{seg}$, $\|\cdot\|_1$ denotes the element-wise $\ell_1$ norm, and $\#M^{in}$ denotes the number of non-zero pixels in $M^{in}$.

\textbf{L1 Loss.} A pixel-wise $\ell_1$ loss normalised by the transparent mask area:
\begin{equation}
    \mathcal{L}_{\mathrm{L1}} = \frac{\left\|(I^{pred} - I^{op}) \odot M^{seg}\right\|_1}{\#M^{seg}},
\end{equation}
where $\|\cdot\|_1$ denotes the element-wise $\ell_1$ norm and $\#M^{seg}$ denotes the number of non-zero pixels in $M^{seg}$.
%%%%%%%%%%%%%%%%%%%%%%%%%%%%%%%%%%%%%%%%%%%%%%%%%%%%%%%%%%%%
% Section B: Mask Refinement Module
%%%%%%%%%%%%%%%%%%%%%%%%%%%%%%%%%%%%%%%%%%%%%%%%%%%%%%%%%%%%

\section{Mask Refinement Module: Architecture and Training}
\label{sec:supp-mrm}

\subsection{Network Architecture}
The Mask Refinement Module (MRM) is a lightweight pixel-wise regression head. Its input is the channel-wise concatenation of three spatial maps: the opacified image $I^{op} \in \mathbb{R}^{3 \times H \times W}$, the transparent reference $I^{tr} \in \mathbb{R}^{3 \times H \times W}$, and the augmented segmentation mask $M_{aug} \in \mathbb{R}^{1 \times H \times W}$ (an augmented version of $M^{seg}$ produced by the mask augmentation scheme described in the main paper), forming a 7-channel input tensor. The network consists of four convolutional stages:
\begin{align}
    \mathbf{h}_1 &= \mathrm{SiLU}\!\left(\mathrm{GN}_{32}\!\left(\mathrm{Conv}_{3\times3}^{7 \to 64}(\mathbf{x})\right)\right), \\
    \mathbf{h}_2 &= \mathrm{SiLU}\!\left(\mathrm{GN}_{32}\!\left(\mathrm{Conv}_{3\times3}^{64 \to 32}(\mathbf{h}_1)\right)\right), \\
    \mathbf{h}_3 &= \mathrm{SiLU}\!\left(\mathrm{GN}_{16}\!\left(\mathrm{Conv}_{3\times3}^{32 \to 16}(\mathbf{h}_2)\right)\right), \\
    M_{refine} &= \sigma\!\left(\mathrm{Conv}_{1\times1}^{16 \to 1}(\mathbf{h}_3)\right) \in (0,1)^{H \times W},
\end{align}
where $\mathrm{GN}_{G}$ denotes Group Normalisation with $G$ groups, all $3{\times}3$ convolutions use $\mathrm{padding}=1$, and $\sigma$ denotes the sigmoid activation. Here $M_{refine} \in (0,1)^{H \times W}$ is the soft probability map used for training supervision (see Sec.~\ref{sec:supp-mrm-loss}). At inference, $M_{refine}$ is further binarised by retaining only connected components overlapping with the segmentation mask $M^{seg}$, followed by morphological closing and opening, to obtain the final binary blending mask $\hat{M}_{refine} \in \{0,1\}^{H \times W}$ used in the compositing equation of the main paper.

\subsection{Loss Function}
\label{sec:supp-mrm-loss}

The MRM is trained with the loss defined in the main paper, where $M_{refine} \in (0,1)^{H \times W}$ is the soft sigmoid output of the network. The two loss terms expand as:
\begin{align}
    \mathcal{L}_{BCE} &= -\left[M^*\log M_{refine} + (1-M^*)\log(1-M_{refine})\right], \\
    \mathcal{L}_{refine} &= \mathcal{L}_{BCE} + \lambda \sum M_{refine}(1 - M_{refine}),
\end{align}
where $M^*$ is the ground-truth binary object mask and $\lambda = 0.1$. The second term is maximised when $M_{refine} = 0.5$ and vanishes at clean binary predictions, acting as a regulariser that pushes the network towards confident mask boundaries.

\subsection{Training Configuration}

The MRM is trained independently of the opacification stage using the AdamW optimiser with learning rate $1 \times 10^{-4}$, batch size 16, for 400 epochs at $512 \times 512$ resolution.

%%%%%%%%%%%%%%%%%%%%%%%%%%%%%%%%%%%%%%%%%%%%%%%%%%%%%%%%%%%%
% Section C: UniPC Sampler Configuration
%%%%%%%%%%%%%%%%%%%%%%%%%%%%%%%%%%%%%%%%%%%%%%%%%%%%%%%%%%%%

\section{UniPC Sampler Configuration}
\label{sec:supp-unipc}

UniPC is a training-free unified predictor--corrector solver for diffusion ODEs. It operates in the noise-prediction parameterisation, where the denoising UNet $\epsilon_\theta$ predicts the injected noise at each timestep. At each step, a $p$-th order multistep predictor (UniP-$p$) advances the noisy latent using the last $p$ noise estimates:
\begin{multline}
    \tilde{z}_{t_i} = \frac{\alpha_{t_i}}{\alpha_{t_{i-1}}}\tilde{z}_{t_{i-1}} - \sigma_{t_i}(e^{h_i}-1)\epsilon_\theta(\tilde{z}_{t_{i-1}}, t_{i-1}) \\
    - \sigma_{t_i}B(h_i)\sum_{m=1}^{p-1}\frac{a_m}{r_m}D_m,
\end{multline}
where $\tilde{z}_{t_i}$ denotes the predictor estimate of the noisy latent at timestep $t_i$, $h_i = \lambda_{t_i} - \lambda_{t_{i-1}}$ is the step size in the half-log-SNR domain, $D_m = \epsilon_\theta(\tilde{z}_{s_m}, s_m) - \epsilon_\theta(\tilde{z}_{t_{i-1}}, t_{i-1})$ are finite differences of noise estimates at auxiliary timesteps $s_m$, $B(h)$ is a non-zero function of $h$, and $\{a_m\}$ are Taylor-expansion coefficients chosen to cancel low-order error terms. A corrector (UniC-$p$) then refines the prediction by incorporating an additional noise estimate at the predicted state, increasing the order of accuracy by one without an extra network call.

At every denoising step, the UNet receives the 9-channel input tensor
\begin{equation}
    x^{(s)}_{\mathrm{in}} = \bigl[\,z_s \;\|\; \mathcal{E}(I^{tr}) \;\|\; \mathcal{D}_{down}(M^{seg})\,\bigr] \in \mathbb{R}^{9 \times h \times w},
\end{equation}
where $\mathcal{E}(I^{tr})$ and $\mathcal{D}_{down}(M^{seg})$ remain fixed throughout the trajectory. We use $S=10$ steps with order-3 multistep updates and uniform timestep spacing. The final step uses a reduced order to improve numerical stability when insufficient history is available for high-order extrapolation.

\section{Generative Opacification: Training Details}
\label{sec:supp-opacification}

\subsection{Noise Schedule and Forward Diffusion}

We adopt a discrete-time DDPM formulation with a linear variance schedule over $T = 1000$ timesteps. The per-step noise variance is defined by linear interpolation between $\beta_{\min}$ and $\beta_{\max}$:
\begin{multline}
    \beta_t = \left(\sqrt{\beta_{\min}} + \frac{t-1}{T-1}\left(\sqrt{\beta_{\max}} - \sqrt{\beta_{\min}}\right)\right)^2, \\
    t = 1, \ldots, T,
\end{multline}
with $\beta_{\min} = 8.5 \times 10^{-4}$ and $\beta_{\max} = 1.2 \times 10^{-2}$. The cumulative signal retention is $\bar{\alpha}_t = \prod_{s=1}^{t}(1 - \beta_s)$. Given the clean latent $z_0 = \mathcal{E}(I^{op})$ produced by the frozen VAE encoder $\mathcal{E}$, the forward process is
\begin{equation}
    q(z_t \mid z_0) = \mathcal{N}\!\left(\sqrt{\bar{\alpha}_t}\,z_0,\,(1 - \bar{\alpha}_t)\,\mathbf{I}\right),
\end{equation}
enabling direct sampling $z_t = \sqrt{\bar{\alpha}_t}\,z_0 + \sqrt{1 - \bar{\alpha}_t}\,\boldsymbol{\epsilon}$, $\boldsymbol{\epsilon} \sim \mathcal{N}(\mathbf{0}, \mathbf{I})$. The learning rate is warmed up linearly over the first 200 iterations before being held at the base learning rate.

%%%%%%%%%%%%%%%%%%%%%%%%%%%%%%%%%%%%%%%%%%%%%%%%%%%%%%%%%%%%
% Section A: Generative Opacification
%%%%%%%%%%%%%%%%%%%%%%%%%%%%%%%%%%%%%%%%%%%%%%%%%%%%%%%%%%%%

\subsection{UNet Architecture}

The denoising backbone is a spatially conditioned UNet whose input channel count is expanded to accommodate the inpainting context. Concretely, the noisy latent $z_t \in \mathbb{R}^{4 \times h \times w}$ is concatenated channel-wise with the VAE-encoded transparent reference $\mathcal{E}(I^{tr}) \in \mathbb{R}^{4 \times h \times w}$ and the spatially downsampled segmentation mask $\mathcal{D}_{down}(M^{seg}) \in \mathbb{R}^{1 \times h \times w}$, yielding a 9-channel input tensor. The key architectural hyperparameters are: base channels 320, channel multipliers $[1,2,4,4]$, attention resolutions $[4,2,1]$, 2 residual blocks per resolution, and 8 attention heads. Each spatial transformer block attends over the CLIP conditioning token $c \in \mathbb{R}^{B \times 1 \times 768}$ via cross-attention with context dimension $d_{\mathrm{ctx}} = 768$.

\subsection{CLIP Image Conditioning}

We condition the UNet on the semantic appearance of $I^{tr}$ using a CLIP ViT-L/14 image encoder. The backbone is kept fully frozen; only the transformer mapper, LayerNorm, and linear projection are trainable. The conditioning pipeline proceeds as follows:
\begin{enumerate}[label=(\roman*)]
    \item \textbf{Backbone encoding.} The frozen CLIP ViT-L/14 backbone processes $I^{tr}$ and returns the pooler output (CLS token), $\mathbf{f} \in \mathbb{R}^{B \times 1024}$.
    \item \textbf{Token reshaping.} $\mathbf{f}$ is reshaped to $\mathbb{R}^{B \times 1 \times 1024}$ to form a single-token sequence.
    \item \textbf{Transformer mapper.} A lightweight Transformer with 1 token position, hidden dimension 1024, 5 layers, and 1 attention head refines the representation, producing $\tilde{\mathbf{f}} \in \mathbb{R}^{B \times 1 \times 1024}$.
    \item \textbf{Normalisation and projection.} $\tilde{\mathbf{f}}$ passes through a LayerNorm and a linear projection $\mathbf{W} \in \mathbb{R}^{1024 \times 768}$, yielding the cross-attention context $c \in \mathbb{R}^{B \times 1 \times 768}$.
\end{enumerate}

\subsection{Perceptual Loss Gating}
\label{sec:supp-lpips}

The LPIPS perceptual loss is applied only at low noise levels, where the UNet output constitutes a faithful image estimate. Concretely, we implement this as a per-sample binary gate
\begin{equation}
    g_i = \mathbf{1}\!\left[t_i < \tau T\right], \quad \tau = 0.3,
\end{equation}
so that perceptual supervision is active only when $t_i < 0.3T$. At higher noise levels ($t_i \geq \tau T$), the decoded prediction deviates significantly from the clean image and perceptual supervision would interfere with the denoising objective. The full training objective is
\begin{equation}
    \mathcal{L} = \mathcal{L}_{\mathrm{LDM}} + \lambda\,\mathcal{L}_{\mathrm{LPIPS}}, \quad \lambda = 0.05.
\end{equation}

\subsection{Mask Augmentation for Conditional Training}

To bridge the gap between ground-truth segmentation masks available during training and the imperfect masks encountered at test time, we apply a geometric augmentation scheme to $M^{seg}$ before using it as the UNet conditioning channel. Specifically, we randomly add or remove small circular and rectangular shapes along the object boundary. Concretely, $n_{\mathrm{shapes}} \in \{3, 4, 5\}$ geometric primitives with side lengths sampled at $r \in [0.05,\,0.07]$ of the image width are placed near the mask contour, followed by a morphological opening with a $3{\times}3$ structuring element and Gaussian smoothing ($\sigma=1.0$) before binarisation. This produces augmented masks that are locally inconsistent with the true boundary while preserving overall shape, encouraging the model to tolerate imperfect segmentation inputs.

\end{document}